\documentclass[letterpaper, 10 pt, journal, twoside]{IEEEtran}
\ifCLASSINFOpdf
\else
\fi
\usepackage{color}
\usepackage{transparent}
\usepackage{import}
\usepackage{multirow}
\usepackage{balance}
\usepackage{graphicx,xcolor}
\usepackage{psfrag}
\usepackage{amsmath,amsfonts} 

\usepackage{amssymb}  
\usepackage{makecell}
\usepackage{hyperref}

\usepackage{booktabs,tabularx,caption,ragged2e}
\usepackage{cite}
\usepackage{times,xspace}
\usepackage[font={footnotesize}]{caption}
\usepackage{makecell}
\usepackage{algorithm} 
\usepackage{algpseudocode} 
\usepackage{nicefrac}
\usepackage{romannum}
\usepackage{subcaption}
\usepackage{siunitx}
\usepackage{url}
\usepackage{array,multirow}
\usepackage[bottom]{footmisc}

\usepackage[makeroom]{cancel}

\usepackage{mathtools}

\usepackage{array}
\usepackage{stmaryrd}
\usepackage{accents}
\usepackage{ntheorem}
\theoremstyle{break}
\theoremheaderfont{\normalfont\bfseries}
\theoremseparator{:}

\newtheorem{definition}{Definition}

\newtheorem{property}{Property}
\newtheorem{theorem}{Theorem}
\RequirePackage[                    
strict=true,                    
style=german                    
]{csquotes}
\usepackage{tikz} 
\newcommand\copyrighttext{%
	\footnotesize \copyright 2023 IEEE. Personal use of this material is permitted. Permission from IEEE must be obtained for all other uses, in any current or future media, including reprinting/republishing this material for advertising or promotional purposes, creating new collective works, for resale or redistribution to servers or lists, or reuse of any copyrighted component of this work in other works.}
\newcommand\copyrightnotice{%
	\begin{tikzpicture}[remember picture,overlay]
		\node[anchor=south,yshift=5pt] at (current page.south) {\fbox{\parbox{\dimexpr\textwidth-\fboxsep-\fboxrule\relax}{\copyrighttext}}};
	\end{tikzpicture}%
}
\definecolor{RPTHred}{RGB}{227, 27, 35}
\definecolor{TUMblue}{RGB}{0, 101, 189}
\usepackage{romannum}

\usepackage{pifont}
\newcommand{\cmark}{\ding{51}}%
\newcommand{\xmark}{\ding{55}}%

\usepackage[symbols, toc,nonumberlist, sanitizesort]{glossaries}
\usepackage{color}
\usepackage{seqsplit}
\usepackage{xargs}
\usepackage{xstring}

\newglossary[glignoredl]{ignored}{glignored}{glignoredin}{Ignored Glossary}

\newglossary{time}{gls1}{glo1}{Time}
\newglossary{longitudinalplanning}{gls2}{glo2}{Longitudinal Trajectory Planning}
\newglossary{lateralplanning}{gls3}{glo3}{Lateral Trajectory Planning}
\newglossary{cosy}{gls4}{glo4}{Coordinate Systems}
\newglossary{misc}{gls5}{glo5}{Miscellaneous}
\newglossary{state}{gls6}{glo6}{State Variables of the Ego Vehicle}
\newglossary{refpath}{gls7}{glo7}{Reference Path}
\newglossary{shape}{gls8}{glo8}{Ego Vehicle Shape}
\newglossary{obs}{gls9}{glo9}{Obstacles}
\newglossary{reach}{gls10}{glo10}{Reachability Analysis}
\newglossary{drivingcorridor}{gls11}{glo11}{Driving Corridor Identification}
\newglossary{constraints}{gls12}{glo12}{Collision Avoidance Constraints}
\newglossary{trajectoryrepairing}{gls13}{glo13}{Trajectory Repairing}
\newglossary{signaltemporallogic}{gls14}{glo14}{Signal Temporal Logic}

\makenoidxglossaries
\glsnoexpandfields

\newcommand{\RobName}{model predictive\xspace}
\newcommand{\MP}{\mathtt{MP}}

\newcommand{\shorteq}{%
	\settowidth{\@tempdima}{-}%
	\resizebox{\@tempdima}{\height}{=}%
}

\newcommand{\ndec}{\glslink{ndec}{\alpha}}
\newglossaryentry{ndec}{
	name={\ensuremath{\ndec}},
	description={Number of tactical decisions},
	sort=5 a,
	type=misc
}

\newcommand{\nobs}{\glslink{nobs}{\sigma}}
\newglossaryentry{nobs}{
	name={\ensuremath{\nobs}},
	description={Number of obstacles},
	sort=5 b,
	type=misc
}

\newcommand{\cartesian}{\mathtt{G}}
\newcommand{\local}{\mathtt{L}}
\newcommand{\vehiclefixed}{\mathtt{V}}
\newcommandx{\cosy}[1]{F^{#1}}
\newcommand{\transformation}{T}

\newcommand{\curvframe}{\glslink{curvframe}{\cosy{\local}}}
\newglossaryentry{curvframe}{
	name={\ensuremath{\curvframe}},
	description={Local, curvilinear coordinate frame},
	sort=b,
	type=cosy
}

\newcommandx{\localvar}[1]{\glslink{localvar}{\leftidx{^{\local}}{#1}}}
\newglossaryentry{localvar}{
	name={\ensuremath{\localvar{\square}}},
	description={Variable $\square$ in $\curvframe$},
	parent=curvframe,
	sort=b,
	type=cosy
}

\newcommand{\cartesianframe}{\glslink{cartesianframe}{\cosy{\cartesian}}}
\newglossaryentry{cartesianframe}{
	name={\ensuremath{\cartesianframe}},
	description={Global, Cartesian coordinate frame},
	sort=b,
	type=cosy
}

\newcommandx{\globalvar}[1]{\glslink{globalvar}{\leftidx{^{\cartesian}}{#1}}}
\newglossaryentry{globalvar}{
	name={\ensuremath{\globalvar{\square}}},
	description={Variable $\square$ in $\cartesianframe$},
	parent=cartesianframe,
	sort=b,
	type=cosy
}

\newcommand{\vehicleframe}{\glslink{vehicleframe}{\cosy{\vehiclefixed}}}
\newglossaryentry{vehicleframe}{
	name={\ensuremath{\vehicleframe}},
	description={Vehicle-fixed coordinate frame},
	sort=b,
	type=cosy
}

\newcommand{\curvframelondir}{\glslink{curvframelondir}{\zeta}}
\newglossaryentry{curvframelondir}{
	name={\ensuremath{\curvframelondir}},
	description={Longitudinal direction in $\curvframe$},
	parent=curvframe,
	sort=b,
	type=cosy
}

\newcommand{\curvframelatdir}{\glslink{curvframelatdir}{\eta}}
\newglossaryentry{curvframelatdir}{
	name={\ensuremath{\curvframelatdir}},
	description={Lateral direction in $\curvframe$},
	parent=curvframe,
	sort=b,
	type=cosy
}

\newcommandx{\transformationcurvcartesian}[1][1={\slon}, usedefault]{\glslink{transformationcurvcartesian}{\transformation_{\local}^{\cartesian}\left(#1\right)}}
\newglossaryentry{transformationcurvcartesian}{
	name={\ensuremath{\transformationcurvcartesian}},
	description={Transformation from $\curvframe$ to $\cartesianframe$},
	sort=b b,
	type=cosy
}

\newcommandx{\transformationvehiclecartesian}[1][1=\orientation, usedefault]{\glslink{transformationvehiclecartesian}{\transformation_{\vehiclefixed}^{\cartesian}\left(#1\right)}}
\newglossaryentry{transformationvehiclecartesian}{
	name={\ensuremath{\transformationvehiclecartesian}},
	description={Transformation from $\vehicleframe$ to $\cartesianframe$},
	sort=b b,
	type=cosy
}

\newcommand{\maxval}[1]{\glslink{maxval}{\overline{#1}}}
\newglossaryentry{maxval}{
	name={\ensuremath{\maxval{\square}}},
	description={Maximum admissible value of a variable $\square$},
	sort=6a,
	type=state
}

\newcommand{\minval}[1]{\glslink{minval}{\underline{#1}}}
\newglossaryentry{minval}{
	name={\ensuremath{\minval{\square}}},
	description={Minimum admissible value of a variable $\square$},
	sort=6b,
	type=state
}

\newcommand{\pos}{s}
\newcommand{\velocity}{v}
\newcommand{\acceleration}{a}
\newcommand{\jerk}{j}

\newcommand{\slon}{\glslink{slon}{\pos_{\curvframelondir}}}
\newglossaryentry{slon}{
	name={\ensuremath{\slon}},
	description={Longitudinal position in $\curvframe$},
	sort=6c,
	type=state
}
\newcommandx{\slonk}[1][1=\timestep, usedefault]{
	\glslink{slonk}{\pos_{{\curvframelondir, #1}}}}
\newglossaryentry{slonk}{
	name={\ensuremath{\slonk}},
	description={Longitudinal position $\slon$ at time step $\timestep$},
	sort=6c,
	parent=slon,
	type=state
}

\newcommand{\dslon}{\glslink{dslon}{\dot{\pos}_{\curvframelondir}}}
\newglossaryentry{dslon}{
	name={\ensuremath{\dslon}},
	description={Time derivative of longitudinal position in $\curvframe$},
	type=ignored,
}

\newcommand{\sminlon}{\glslink{sminlon}{\minval{\pos}_{\curvframelondir}}}
\newglossaryentry{sminlon}{
	name={\ensuremath{\sminlon}},
	description={Minimum value of $\slon$},
	sort=6c,
	type=state,
	parent=slon
}

\newcommandx{\sminlonk}[1][1=\timestep, usedefault]{\glslink{sminlonk}{\minval{\pos}_{\curvframelondir, #1}}}
\newglossaryentry{sminlonk}{
	name={\ensuremath{\sminlonk}},
	description={Minimum value of $\slon$ at time step $\timestep$},
	sort=6c,
	type=state,
	parent=sminlon
}

\newcommand{\smaxlon}{\glslink{smaxlon}{\maxval{\pos}_{\curvframelondir}}}
\newglossaryentry{smaxlon}{
	name={\ensuremath{\smaxlon}},
	description={Maximum value of $\slon$},
	sort=6c,
	type=state,
	parent=slon
}

\newcommandx{\smaxlonk}[1][1=\timestep, usedefault]{\glslink{smaxlonk}{\maxval{\pos}_{\curvframelondir, #1}}}
\newglossaryentry{smaxlonk}{
	name={\ensuremath{\smaxlonk}},
	description={Maximum value of $\slon$ at time step $\timestep$},
	sort=6c,
	type=state,
	parent=smaxlon
}

\newcommand{\vlon}{\glslink{vlon}{\velocity_{\curvframelondir}}}
\newglossaryentry{vlon}{
	name={\ensuremath{\vlon}},
	description={Longitudinal velocity in $\curvframe$},
	sort=6d,
	type=state
}
\newcommandx{\vlonk}[1][1=\timestep, usedefault]{\glslink{vlonk}{\velocity_{\curvframelondir, #1}}}
\newglossaryentry{vlonk}{
	name={\ensuremath{\vlonk}},
	description={Longitudinal velocity $\vlon$ at time step $\timestep$},
	sort=6d,
	type=state,
	parent=vlon
}

\newcommand{\vminlon}{\glslink{vminlon}{\minval{\velocity}_{\curvframelondir}}}
\newglossaryentry{vminlon}{
	name={\ensuremath{\vminlon}},
	description={Minimum value of $\vlon$},
	sort=6d,
	type=state,
	parent=vlon
}

\newcommand{\vmaxlon}{\glslink{vmaxlon}{\maxval{\velocity}_{\curvframelondir}}}
\newglossaryentry{vmaxlon}{
	name={\ensuremath{\vmaxlon}},
	description={Maximum value of $\vlon$},
	sort=6d,
	type=state,
	parent=vlon
}

\newcommand{\alon}{\glslink{alon}{\acceleration_{\curvframelondir}}}
\newglossaryentry{alon}{
	name={\ensuremath{\alon}},
	description={Longitudinal acceleration in $\curvframe$},
	sort=6e,
	type=state,
}

\newcommandx{\alonk}[1][1=\timestep, usedefault]{\acceleration_{\curvframelondir, #1}}
\newglossaryentry{alonk}{
	name={\ensuremath{\alonk}},
	description={Longitudinal acceleration $\alon$ at time step $\timestep$},
	sort=6e,
	type=state,
	parent=alon
}

\newcommand{\aminlon}{\glslink{aminlon}{\minval{\acceleration}_{\curvframelondir}}}
\newglossaryentry{aminlon}{
	name={\ensuremath{\aminlon}},
	description={Minimum value of $\alon$},
	sort=6e,
	type=state,
	parent=alon
}

\newcommand{\amaxlon}{\glslink{amaxlon}{\maxval{\acceleration}_{\curvframelondir}}}
\newglossaryentry{amaxlon}{
	name={\ensuremath{\amaxlon}},
	description={Maximum value of $\alon$},
	sort=6e,
	type=state,
	parent=alon
}

\newcommand{\jlon}{\glslink{jlon}{\jerk_{\curvframelondir}}}
\newglossaryentry{jlon}{
	name={\ensuremath{\jlon}},
	description={Longitudinal jerk in $\curvframe$},
	sort=6f,
	type=state,
}

\newcommandx{\jlonk}[1][1=\timestep, usedefault]{\glslink{jlonk}{\jerk_{\curvframelondir, #1}}}
\newglossaryentry{jlonk}{
	name={\ensuremath{\jlonk}},
	description={Longitudinal jerk $jlonk$ at time step $\timestep$},
	sort=6f,
	type=state,
	parent=jlon
}

\newcommand{\slat}{\glslink{slat}{\pos_{\curvframelatdir}}}
\newglossaryentry{slat}{
	name={\ensuremath{\slat}},
	description={Lateral position in $\curvframe$},
	sort=6g,
	type=state
}

\newcommandx{\slatk}[1][1=\timestep, usedefault]{
	\glslink{slatk}{\pos_{{\curvframelatdir, #1}}}}
\newglossaryentry{slatk}{
	name={\ensuremath{\slatk}},
	description={Lateral position $\slat$ at time step $\timestep$},
	sort=6g,
	type=state,
	parent=slat
}

\newcommand{\dslat}{\glslink{dslat}{\dot{\pos}_{\curvframelatdir}}}
\newglossaryentry{dslat}{
	name={\ensuremath{\dslat}},
	description={Time derivative of lateral position in $\curvframe$},
	type=ignored
}

\newcommand{\vlat}{\glslink{vlat}{\velocity_{\curvframelatdir}}}
\newglossaryentry{vlat}{
	name={\ensuremath{\vlat}},
	description={Lateral velocity in $\curvframe$},
	sort=6h,
	type=state
}

\newcommandx{\vlatk}[1][1=\timestep, usedefault]{\glslink{vlatk}{\velocity_{\curvframelatdir, #1}}}
\newglossaryentry{vlatk}{
	name={\ensuremath{\vlatk}},
	description={Lateral velocity $\vlat$ at time step $\timestep$},
	sort=6h,
	type=state,
	parent=vlat
}

\newcommand{\vminlat}{\glslink{vminlat}{\minval{\velocity}_{\curvframelatdir}}}
\newglossaryentry{vminlat}{
	name={\ensuremath{\vminlat}},
	description={Minimum value of $\vlat$},
	sort=6h,
	type=state,
	parent=vlat
}

\newcommand{\vmaxlat}{\glslink{vmaxlat}{\maxval{\velocity}_{\curvframelatdir}}}
\newglossaryentry{vmaxlat}{
	name={\ensuremath{\vmaxlat}},
	description={Minimum value of $\vlat$},
	sort=6h,
	type=state,
	parent=vlat
}

\newcommand{\alat}{\glslink{alat}{\acceleration_{\curvframelatdir}}}
\newglossaryentry{alat}{
	name={\ensuremath{\alat}},
	description={Lateral acceleration in $\curvframe$},
	sort=6i,
	type=state
}

\newcommandx{\alatk}[1][1=\timestep, usedefault]{\acceleration_{\curvframelatdir, #1}}
\newglossaryentry{alatk}{
	name={\ensuremath{\alatk}},
	description={Lateral acceleration $\alatk$ at time step $\timestep$},
	sort=6i,
	type=state,
	parent=alat
}

\newcommand{\aminlat}{\glslink{aminlat}{\minval{\acceleration}_{\curvframelatdir}}}
\newglossaryentry{aminlat}{
	name={\ensuremath{\aminlat}},
	description={Minimum value of $\alat$},
	sort=6i,
	type=state,
	parent=alat
}

\newcommand{\amaxlat}{\glslink{amaxlat}{\maxval{\acceleration}_{\curvframelatdir}}}
\newglossaryentry{amaxlat}{
	name={\ensuremath{\amaxlat}},
	description={Maximum value of $\alat$},
	sort=6i,
	type=state,
	parent=alat
}

\newcommand{\sx}{\glslink{sx}{\pos_{x}}}
\newglossaryentry{sx}{
	name={\ensuremath{\sx}},
	description={Longitudinal position in $\cartesianframe$},
	sort=6j,
	type=state
}

\newcommand{\sy}{\glslink{sy}{\pos_{y}}}
\newglossaryentry{sy}{
	name={\ensuremath{\sy}},
	description={Lateral position in $\cartesianframe$},
	sort=6k,
	type=state
}

\newcommand{\orientation}{\glslink{orientation}{\theta}}
\newglossaryentry{orientation}{
	name={\ensuremath{\orientation}},
	description={Orientation in $\cartesianframe$},
	sort=6l,
	type=state
}

\newcommandx{\orientationk}[1][1=\timestep, usedefault]{\glslink{orientationk}{\theta_{#1}}}
\newglossaryentry{orientationk}{
	name={\ensuremath{\orientationk}},
	description={Orientation $\orientation$ at time step $\timestep$},
	sort=6l,
	type=state,
	parent=orientation
}

\newcommand{\curvature}{\glslink{curvature}{\kappa}}
\newglossaryentry{curvature}{
	name={\ensuremath{\curvature}},
	description={Curvature in $\cartesianframe$},
	sort=6l,
	type=state
}

\newcommandx{\curvaturek}[1][1=\timestep, usedefault]{\curvature_{#1}}
\newglossaryentry{curvaturek}{
	name={\ensuremath{\curvaturek}},
	description={Curvature $\curvature$ at time step $\timestep$},
	sort=6l,
	type=state,
	parent=curvature
}

\newcommand{\refpathsym}{\Gamma}

\newcommandx{\refpath}[1][1=\slon, usedefault]{\glslink{refpath}{\refpathsym(#1)}}
\newglossaryentry{refpath}{
	name={\ensuremath{\refpath}},
	description={Reference path},
	sort=7a,
	type=refpath
}

\newcommand{\reforientation}{\glslink{reforientation}{\orientation_\refpathsym}}
\newglossaryentry{reforientation}{
	name={\ensuremath{\reforientation}},
	description={Orientation of reference path},
	sort=7b,
	type=refpath
}

\newcommand{\refcurvature}{\glslink{refcurvature}{\curvature_\refpathsym}}
\newglossaryentry{refcurvature}{
	name={\ensuremath{\refcurvature}},
	description={Curvature of reference path},
	sort=7c,
	type=refpath
}

\newcommand{\minrefcurvature}{\glslink{minrefcurvature}{\minval{\curvature}_\refpathsym}}
\newglossaryentry{minrefcurvature}{
	name={\ensuremath{\minrefcurvature}},
	description={Minimum curvature of reference path},
	sort=7d,
	type=refpath
}

\newcommand{\maxrefcurvature}{\glslink{maxrefcurvature}{\maxval{\curvature}_\refpathsym}}
\newglossaryentry{maxrefcurvature}{
	name={\ensuremath{\maxrefcurvature}},
	description={Maximum curvature of reference path},
	sort=7f,
	type=refpath
}

\newcommand{\idxcentercircle}{i}
\newcommand{\centerpoint}{c}

\newcommand{\wheelbase}{\glslink{wheelbase}{\ell}}
\newglossaryentry{wheelbase}{
	name={\ensuremath{\wheelbase}},
	description={Wheelbase},
	sort=8a,
	type=shape
}

\newcommand{\radius}{\glslink{radius}{r}}
\newglossaryentry{radius}{
	name={\ensuremath{\radius}},
	description={Radius of the circles approximating the vehicle shape},
	sort=8b,
	type=shape
}

\newcommandx{\centercircle}[2][1=\idxcentercircle, 2=\timestep, usedefault]{\glslink{centercircle}{\centerpoint^{(#1)}_{#2}}\glslink{centercirclek}{}}
\newglossaryentry{centercircle}{
	name={\ensuremath{\centercircle[][ ]}},
	description={Center of the $\idxcentercircle$-th circle approximating the vehicle shape},
	sort=8c,
	type=shape
}
\newglossaryentry{centercirclek}{
	name={\ensuremath{\centercircle}},
	description={Center $\centercircle$ at time step $\timestep$},
	sort=8c,
	type=shape,
	parent=centercircle,
}

\newcommandx{\globalcentercirclek}[2][1=\idxcentercircle, 2=\timestep, usedefault]{\glslink{globalcentercirclek}{\globalvar{\centerpoint}^{(#1)}_{#2}}}
\newglossaryentry{globalcentercirclek}{
	name={\ensuremath{\globalcentercirclek}},
	description={Center $\centercircle$ in $\cartesianframe$ at time step $\timestep$},
	sort=8c,
	type=shape,
	parent=centercircle
}

\newcommandx{\globalcentercirclerayk}[2][1=\idxcentercircle, 2=\timestep, usedefault]{\glslink{globalcentercirclerayk}{\globalvar{\hat{\centerpoint}}^{(#1)}_{#2}}}
\newglossaryentry{globalcentercirclerayk}{
	name={\ensuremath{\globalcentercirclerayk}},
	description={Center $\centercircle$ in $\cartesianframe$ at time step $\timestep$},
	sort=8c,
	type=shape,
	parent=centercircle
}

\newcommandx{\localcentercirclek}[2][1=\idxcentercircle, 2=\timestep, usedefault]{\glslink{localcentercirclek}{\localvar{\centerpoint}^{(#1)}_{#2}}}
\newglossaryentry{localcentercirclek}{
	name={\ensuremath{\localcentercirclek}},
	description={Center $\centercircle$ in $\curvframe$ at time step $\timestep$},
	sort=8c,
	type=shape,
	parent=centercircle
}

\newcommand{\cut}{\mathrm{cut}}

\newcommand{\conttime}{\glslink{time}{t}}
\newglossaryentry{time}{
	name={\ensuremath{\conttime}},
	description={Time},
	sort=1 a,
	type=time,
}

\newcommand{\dt}{\glslink{dt}{\Delta \conttime}}
\newglossaryentry{dt}{
	name={\ensuremath{\dt}},
	description={Time increment},
	sort=1 b,
	type=time,
}

\newcommand{\thorizon}{\glslink{thorizon}{h}}
\newglossaryentry{thorizon}{
	name={\ensuremath{\thorizon}},
	description={Final time step},
	sort=1 c,
	type=time
}

\newcommand{\tinit}{\glslink{tinit}{0}}
\newglossaryentry{tinit}{
	name={\ensuremath{\tinit}},
	description={Initial time step},
	sort=1 d,
	type=time
}

\newcommand{\tcutoff}{\glslink{tinit}{\cut}}
\newglossaryentry{tcutoff}{
	name={\ensuremath{\tcutoff}},
	description={Cut-off time step},
	sort=1 h,
	type=time
}

\newcommand{\timestep}{\glslink{timestep_b}{k}}
\newglossaryentry{timestep_a}{
	name=\ensuremath{\timestep},
	description={Discrete time},
	sort=1 f,
	type=time,
}
\newglossaryentry{timestep_b}{
	name={\ensuremath{\square_k}},
	description={Variable $\square$ at time step $\timestep$},
	parent=timestep_a,
	sort=1 g,
	type=time,
}

\newcommand{\lon}{\mathrm{lon}}
\newcommand{\lat}{\mathrm{lat}}
\newcommand{\R}{\mathcal{R}}

\newcommand{\systemmatrix}{\glslink{systemmatrix}{A}}
\newglossaryentry{systemmatrix}{
	name={\ensuremath{\systemmatrix}},
	description={System matrix},
	sort=c e
}

\newcommand{\inputmatrix}{\glslink{inputmatrix}{B}}
\newglossaryentry{inputmatrix}{
	name={\ensuremath{\inputmatrix}},
	description={Input matrix},
	sort=c f
}

\newcommand{\x}{\glslink{x}{x}}
\newglossaryentry{x}{
	name={\ensuremath{\x}},
	description={State},
	sort=c g
}

\newcommand{\systeminput}{\glslink{systeminput}{u}}
\newglossaryentry{systeminput}{
	name={\ensuremath{\systeminput}},
	description={Input},
	sort=c h
}

\newcommand{\X}{\glslink{X}{\mathcal{X}}}
\newglossaryentry{X}{
	name={\ensuremath{\X}},
	description={Admissible set of states},
	sort=c i
}

\newcommand{\U}{\glslink{U}{\mathcal{U}}}
\newglossaryentry{U}{
	name={\ensuremath{\U}},
	description={Admissible set of inputs},
	sort=c j
}

\newcommandx{\initialstatetrajectory}{\glslink{initialstatetrajectory}{\chi^{\mathtt{int}}}}
\newglossaryentry{initialstatetrajectory}{
	name={\ensuremath{\initialstatetrajectory[ ]}},
	description={State trajectory},
	sort=c k
}

\newcommandx{\repairedstatetrajectory}{\glslink{repairedstatetrajectory}{\chi^{\mathtt{rep}}}}
\newglossaryentry{repairedstatetrajectory}{
	name={\ensuremath{\repairedstatetrajectory[ ]}},
	description={State trajectory},
	sort=c k
}

\newcommandx{\initialinputtrajectory}{\glslink{initialinputtrajectory}{\chi^{\mathtt{int}}}}
\newglossaryentry{initialinputtrajectory}{
	name={\ensuremath{\initialinputtrajectory[ ]}},
	description={State trajectory},
	sort=c k
}

\newcommandx{\repairedinputtrajectory}{\glslink{repairedinputtrajectory}{\chi^{\mathtt{rep}}}}
\newglossaryentry{repairedinputtrajectory}{
	name={\ensuremath{\repairedinputtrajectory[ ]}},
	description={State trajectory},
	sort=c k
}

\newcommandx{\statetrajectory}[1][1=i, usedefault]{\glslink{statetrajectory}{X_{#1}}}
\newglossaryentry{statetrajectory}{
	name={\ensuremath{\statetrajectory[ ]}},
	description={State trajectory},
	sort=c k
}

\newcommandx{\inputtrajectory}[1][1=i, usedefault]{\glslink{inputtrajectory}{U_{#1}}}
\newglossaryentry{inputtrajectory}{
	name={\ensuremath{\inputtrajectory[ ]}},
	description={Input trajectory},
	sort=c l
}

\newcommand{\utotal}{\glslink{utotal}{\systeminput}}
\newglossaryentry{utotal}{
	name={\ensuremath{\utotal}},
	description={Input},
	type=trajectoryrepairing,
	sort=2 e
}

\newcommand{\xtotal}{\glslink{xtotal}{\x}}
\newglossaryentry{xtotal}{
	name={\ensuremath{\xtotal}},
	description={State},
	type=trajectoryrepairing,
	sort=2 j
}

\newcommand{\costtotal}{\glslink{costtotal}{J}}
\newglossaryentry{costtotal}{
	name={\ensuremath{\costtotal}},
	description={Cost function},
	type=trajectoryrepairing,
	sort=2 i
}

\newcommand{\robusttotal}{\glslink{robosttotal}{P}}
\newglossaryentry{robosttotal}{
	name={\ensuremath{\robusttotal}},
	description={Robustness function},
	type=trajectoryrepairing,
	sort=2 h
}

\newcommand{\xinit}{\glslink{xinit}{\x_0}}
\newglossaryentry{xinit}{
	name={\ensuremath{\xinit}},
	description={Initial trajectory},
	type=trajectoryrepairing,
	sort=2 j
}

\newcommandx{\xtotalk}[1][1=\timestep, usedefault]{\glslink{xtotalk}{\xtotal_{{#1}}}}
\newglossaryentry{xtotalk}{
	name={\ensuremath{\xtotalk}},
	description={State at time step $\timestep$},
	parent=xtotal,
	type=trajectoryrepairing,
	sort=2 m
}

\newcommandx{\xrepairk}[1][1=\timestep, usedefault]{\glslink{xrepairk}{\xtotal^{\mathtt{rep}}_{{#1}}}}
\newglossaryentry{xrepairk}{
	name={\ensuremath{\xrepairk}},
	description={State at time step $\timestep$ of the repaired trajectory},
	parent=xtotal,
	type=trajectoryrepairing,
	sort=2 m
}

\newcommandx{\xinitk}[1][1=\timestep, usedefault]{\glslink{xinitk}{\xtotal^{\mathtt{int}}_{{#1}}}}
\newglossaryentry{xinitk}{
	name={\ensuremath{\xinitk}},
	description={State at time step $\timestep$ of the initial trajectory},
	parent=xtotal,
	type=trajectoryrepairing,
	sort=2 m
}

\newcommandx{\uinitk}[1][1=\timestep, usedefault]{\glslink{uinitk}{\utotal^{\mathtt{int}}_{{#1}}}}
\newglossaryentry{uinitk}{
	name={\ensuremath{\uinitk}},
	description={Input at time step $\timestep$ of the initial trajectory},
	parent=xtotal,
	type=trajectoryrepairing,
	sort=2 m
}

\newcommandx{\utotalk}[1][1=\timestep, usedefault]{\glslink{utotalk}{\utotal_{{#1}}}}
\newglossaryentry{utotalk}{
	name={\ensuremath{\utotalk}},
	description={Input at time step $\timestep$},
	parent=utotal,
	type=trajectoryrepairing,
	sort=2 n
}

\newcommandx{\urepairk}[1][1=\timestep, usedefault]{\glslink{urepairk}{\utotal^{\mathtt{rep}}_{{#1}}}}
\newglossaryentry{urepairk}{
	name={\ensuremath{\urepairk}},
	description={Input at time step $\timestep$ of the repaired trajectory},
	parent=utotal,
	type=trajectoryrepairing,
	sort=2 m
}

\newcommandx{\Xtotalk}[1][1=\timestep, usedefault]{\glslink{Xtotalk}{\X_{#1}}}
\newglossaryentry{Xtotalk}{
	name={\ensuremath{\Xtotalk}},
	description={Set of admissible states at time step $\timestep$},
	type=trajectoryrepairing,
	sort=2 f
}

\newcommandx{\Utotalk}[1][1=\timestep, usedefault]{\glslink{Utotalk}{\U_{#1}}}
\newglossaryentry{Utotalk}{
	name={\ensuremath{\Utotalk}},
	description={Set of admissible inputs at time step $\timestep$},
	type=trajectoryrepairing,
	sort=2 g
}

\newcommand{\reppredicate}{\glslink{reppredicate}{\hat{\psi}}}

\newglossaryentry{reppredicate}{
	name={\ensuremath{\reppredicate}},
	description={State},
	type=trajectoryrepairing,
	sort=2 j
}

\newcommand{\Alon}{\glslink{Alon}{\systemmatrix_{\lon}}}
\newglossaryentry{Alon}{
	name={\ensuremath{\Alon}},
	description={System matrix},
	type=longitudinalplanning,
	sort=2 a
}
\newcommandx{\Alonk}[1][1=\timestep, usedefault]{\glslink{Alonk}{\systemmatrix_{\lon, #1}}}
\newglossaryentry{Alonk}{
	name={\ensuremath{\Alonk}},
	description={System matrix at time step $\timestep$},
	parent=Alon,
	type=longitudinalplanning,
	sort=2 a
}

\newcommandx{\Blon}{\glslink{Blon}{\inputmatrix_{\lon}}}
\newglossaryentry{Blon}{
	name={\ensuremath{\Blon}},
	description={Input matrix trajectory planning},
	type=longitudinalplanning,
	sort=2 b
}
\newcommandx{\Blonk}[1][1=\timestep, usedefault]{\glslink{Blonk}{\inputmatrix_{\lon, #1}}}
\newglossaryentry{Blonk}{
	name={\ensuremath{\Blonk}},
	description={Input matrix at time step $\timestep$},
	parent=Blon,
	type=longitudinalplanning,
	sort=2 b
}

\newcommand{\xlon}{\glslink{xlon}{\x_\lon}}
\newglossaryentry{xlon}{
	name={\ensuremath{\xlon}},
	description={State},
	type=longitudinalplanning,
	sort=2 c
}

\newcommandx{\xlonk}[1][1=\timestep, usedefault]{\glslink{xlonk}{\x_{{\lon, #1}}}}
\newglossaryentry{xlonk}{
	name={\ensuremath{\xlonk}},
	description={State at time step $\timestep$},
	parent=xlon,
	type=longitudinalplanning,
	sort=2 c
}

\newcommand{\ulon}{\glslink{ulon}{\systeminput_\lon}}
\newglossaryentry{ulon}{
	name={\ensuremath{\ulon}},
	description={Input},
	type=longitudinalplanning,
	sort=2 e
}

\newcommandx{\ulonk}[1][1=\timestep, usedefault]{\glslink{ulonk}{\systeminput_{\lon,#1}}}
\newglossaryentry{ulonk}{
	name={\ensuremath{\ulonk}},
	description={Input at time step $\timestep$},
	parent=ulon,
	type=longitudinalplanning,
	sort=2 e
}

\newcommandx{\Xlonk}[1][1=\timestep, usedefault]{\glslink{Xlonk}{\X_{\lon, #1}}}
\newglossaryentry{Xlonk}{
	name={\ensuremath{\Xlonk}},
	description={Set of admissible states at time step $\timestep$},
	type=longitudinalplanning,
	sort=2 f
}

\newcommandx{\Ulonk}[1][1=\timestep, usedefault]{\glslink{Ulonk}{\U_{\lon, #1}}}
\newglossaryentry{Ulonk}{
	name={\ensuremath{\Ulonk}},
	description={Set of admissible inputs at time step $\timestep$},
	type=longitudinalplanning,
	sort=2 g
}

\newcommandx{\statetrajectorylon}{\glslink{statetrajectorylon}{\statetrajectory[\lon]}}
\newglossaryentry{statetrajectorylon}{
	name={\ensuremath{\statetrajectorylon}},
	description={Longitudinal state trajectory},
	type=longitudinalplanning,
	sort=2 h
}

\newcommand{\costlon}{\glslink{costlon}{J_\lon}}
\newglossaryentry{costlon}{
	name={\ensuremath{\costlon}},
	description={Cost function},
	type=longitudinalplanning,
	sort=2 i
}

\newcommand{\Alat}{\glslink{Alat}{\systemmatrix_{\lat}}}
\newglossaryentry{Alat}{
	name={\ensuremath{\Alat}},
	description={System matrix},
	sort=3 a,
	type=lateralplanning,
}
\newcommandx{\Alatk}[1][1=\timestep, usedefault]{\glslink{Alatk}{\systemmatrix_{\lat, #1}}}
\newglossaryentry{Alatk}{
	name={\ensuremath{\Alatk}},
	description={System matrix at time step $\timestep$},
	sort=3 b,
	parent=Alat,
	type=lateralplanning,
}

\newcommandx{\Blat}{\glslink{Blat}{\inputmatrix_{\lat}}}
\newglossaryentry{Blat}{
	name={\ensuremath{\Blat}},
	description={Input matrix},
	sort=3 c,
	type=lateralplanning,
}
\newcommandx{\Blatk}[1][1=\timestep, usedefault]{\glslink{Blatk}{\inputmatrix_{\lat, #1}}}
\newglossaryentry{Blatk}{
	name={\ensuremath{\Blatk}},
	description={Input matrix at time step $\timestep$},
	sort=3 d,
	parent=Blat,
	type=lateralplanning,
}

\newcommand{\xlat}{\glslink{xlat}{\x_\lat}}
\newglossaryentry{xlat}{
	name={\ensuremath{\xlat}},
	description={State},
	sort=3 f,
	type=lateralplanning,
}

\newcommandx{\xlatk}[1][1=\timestep, usedefault]{\glslink{xlatk}{\x_{{\lat, #1}}}}
\newglossaryentry{xlatk}{
	name={\ensuremath{\xlatk}},
	description={State at time step $\timestep$},
	sort=3 g,
	parent=xlat,
	type=lateralplanning,
}

\newcommand{\ulat}{\glslink{ulat}{\systeminput_\lat}}
\newglossaryentry{ulat}{
	name={\ensuremath{\ulat}},
	description={Input},
	sort=3 h,
	type=lateralplanning,
}

\newcommandx{\ulatk}[1][1=\timestep, usedefault]{\glslink{ulatk}{\systeminput_{\lat,#1}}}
\newglossaryentry{ulatk}{
	name={\ensuremath{\ulatk}},
	description={Input at time step $\timestep$},
	sort=3 i,
	type=lateralplanning,
	parent=ulat
}

\newcommandx{\Xlatk}[1][1=\timestep, usedefault]{\glslink{Xlatk}{\X_{\lat, #1}}}
\newglossaryentry{Xlatk}{
	name={\ensuremath{\Xlatk}},
	description={Set of admissible states at time step $\timestep$},
	type=lateralplanning,
	sort=3 j
}

\newcommandx{\Ulatk}[1][1=\timestep, usedefault]{\glslink{Ulatk}{\U_{\lat, #1}}}
\newglossaryentry{Ulatk}{
	name={\ensuremath{\Ulatk}},
	description={Set of admissible inputs at time step $\timestep$},
	type=lateralplanning,
	sort=3 k
}

\newcommandx{\statetrajectorylat}{\glslink{statetrajectorylat}{\statetrajectory[\lat]}}
\newglossaryentry{statetrajectorylat}{
	name={\ensuremath{\statetrajectorylat}},
	description={Lateral state trajectory},
	type=lateralplanning,
	sort=3 l
}

\newcommand{\costlat}{\glslink{costlat}{J_\lat}}
\newglossaryentry{costlat}{
	name={\ensuremath{\costlat}},
	description={Cost function},
	type=lateralplanning,
	sort=3 m
}

\newcommandx{\obstaclesk}[1][1=\globalvar, usedefault]{\glslink{obstaclesk}{#1{\mathcal{O}}_{\timestep}}}
\newglossaryentry{obstaclesk}{
	name={\ensuremath{\obstaclesk}},
	description={Occupancy sets of all obstacles in $\cartesianframe$ at time step $\timestep$},
	type=obs,
	sort=9a
}

\newcommandx{\obstaclescirclek}[1][1=\globalvar, usedefault]{\glslink{obstaclescirclek}{#1{\mathcal{O}}^{\basiccircleshape}_{\timestep}}}
\newglossaryentry{obstaclescirclek}{
	name={\ensuremath{\obstaclescirclek}},
	description={Occupancy sets $\obstaclesk$ dilated with circle $\basiccircleshape$},
	type=obs,
	sort=9c
}

\newcommand{\Areach}{\glslink{Areach}{\systemmatrix_{\R}}}
\newglossaryentry{Areach}{
	name={\ensuremath{\Areach}},
	description={System matrix},
	sort=10a,
	type=reach,
}

\newcommand{\Breach}{\glslink{Breach}{\inputmatrix_{\R}}}
\newglossaryentry{Breach}{
	name={\ensuremath{\Breach}},
	description={Input matrix},
	sort=10b,
	type=reach,
}

\newcommand{\xreach}{\glslink{xreach}{\x_{\R}}}
\newglossaryentry{xreach}{
	name={\ensuremath{\xreach}},
	description={State},
	sort=10c,
	type=reach,
}

\newcommandx{\xreachk}[1][1=\timestep, usedefault]{\glslink{xreachk}{\x_{\R, #1}}}
\newglossaryentry{xreachk}{
	name={\ensuremath{\xreachk}},
	description={State $\xreach$ at time step $\timestep$},
	sort=10c,
	type=reach,
	parent=xreach
}

\newcommand{\ureach}{\glslink{ureach}{\glslink{ureach}{\systeminput_{\R}}}}
\newglossaryentry{ureach}{
	name={\ensuremath{\ureach}},
	description={Input},
	sort=10d,
	type=reach,
}

\newcommandx{\ureachk}[1][1=\timestep, usedefault]{\glslink{ureachk}{\systeminput_{\R, #1}}}
\newglossaryentry{ureachk}{
	name={\ensuremath{\ureachk}},
	description={Input $\ureach$ at time step $\timestep$},
	sort=10d,
	type=reach,
	parent=ureach
}

\newcommandx{\Xreachk}[1][1=\timestep, usedefault]{\glslink{Xreachk}{\X_{\R, #1}}}
\newglossaryentry{Xreachk}{
	name={\ensuremath{\Xreachk}},
	description={Set of admissible states at time step $\timestep$},
	type=reach,
	sort=10e
}

\newcommandx{\Ureachk}[1][1=\timestep, usedefault]{\glslink{Ureachk}{\U_{\R, #1}}}
\newglossaryentry{Ureachk}{
	name={\ensuremath{\Ureachk}},
	description={Set of admissible inputs at time step $\timestep$},
	type=reach,
	sort=10f
}

\newcommandx{\forbiddenset}[1][1=\timestep, usedefault]{\glslink{forbiddenset}{\mathcal{F}_{#1}}}
\newglossaryentry{forbiddenset}{
	name={\ensuremath{\forbiddenset}},
	description={Set of forbidden states at time step $\timestep$},
	type=reach,
	sort=10g
}

\newcommandx{\egooccupancy}[2][1={(\xreachk)}, 2=\globalvar, usedefault]{\glslink{egooccupancy}{#2{\mathcal{Q}}#1}}
\newglossaryentry{egooccupancy}{
	name={\ensuremath{\egooccupancy}},
	description={Occupancy of ego vehicle in $\cartesianframe$ at state $\xreachk$},
	type=reach,
	sort=10h
}

\newcommand{\exact}{\mathtt{e}}
\newcommand{\exactval}[1]{#1^{\exact}}

\newcommandx{\exactreach}[1][1=\timestep, usedefault]{\glslink{exactreach}{\exactval{\R}_{#1}}}
\newglossaryentry{exactreach}{
	name={\ensuremath{\exactreach}},
	description={Exact reachable set at time step $\timestep$},
	type=reach,
	sort=10i
}

\newcommandx{\reachk}[1][1=\timestep, usedefault]{\glslink{reachk}{\R_{#1}}}
\newglossaryentry{reachk}{
	name={\ensuremath{\reachk}},
	description={Approx. reachable set at time step $\timestep$},
	type=reach,
	sort=10j
}

\newcommand{\D}{\mathcal{D}}
\newcommandx{\exactdrivablearea}[1][1=\timestep, usedefault]{\glslink{exactdrivablearea}{\exactval{\D}_{#1}}}
\newglossaryentry{exactdrivablearea}{
	name={\ensuremath{\exactdrivablearea}},
	description={Exact drivable area at time step $\timestep$},
	type=reach,
	sort=10k
}

\newcommandx{\drivableareak}[1][1=\timestep, usedefault]{\glslink{drivableareak}{\D_{#1}}}
\newglossaryentry{drivableareak}{
	name={\ensuremath{\drivableareak}},
	description={Approx. drivable area at time step $\timestep$},
	type=reach,
	sort=10l
}

\newcommand{\graph}{\mathcal{G}}

\newcommand{\graphreachability}{\glslink{graphreachability}\graph_{\R}}
\newglossaryentry{graphreachability}{
	name={\ensuremath{\graphreachability}},
	description={Reachability graph},
	type=reach,
	sort=10m
}

\newcommand{\idxreach}{i}
\newcommand{\baseset}{\R}

\newcommandx{\sBik}[2][1=(\idxreach), 2=\timestep, usedefault]{\glslink{sBik}{\baseset^{#1}_{#2}}}
\newglossaryentry{sBik}{
	name={\ensuremath{\sBik}},
	description={$\idxreach$-th base set at time step $\timestep$},
	type=reach,
	sort=10n
}

\newcommand{\polytope}{\mathcal{P}}
\newcommandx{\sPikl}[3][1=(\idxreach), 2=\timestep, usedefault]{\glslink{sPikllondir}{\polytope^{#1}_{#3, #2}} \glslink{sPikllatdir}{}}
\newglossaryentry{sPikllondir}{
	name={\ensuremath{\sPikl{\curvframelondir}}},
	description={$\idxreach$-th polytope in the $(\slon, \vlon)$ plane at time step $\timestep$},
	type=reach,
	sort=10o
}
\newglossaryentry{sPikllatdir}{
	name={\ensuremath{\sPikl{\curvframelatdir}}},
	description={$\idxreach$-th polytope in the $(\slat, \vlat)$ plane at time step $\timestep$},
	type=reach,
	sort=10p
}

\newcommand{\aabb}{\D}
\newcommandx{\sAik}[2][1=(\idxreach), 2=\timestep, usedefault]{\glslink{sAik}{\aabb^{#1}_{#2}}}
\newglossaryentry{sAik}{
	name={\ensuremath{\sAik}},
	description={Projection of $\sBik$ onto position domain},
	type=reach,
	sort=10q
}

\newcommand{\prop}{\mathtt{prop}}
\newcommandx{\propreachk}[1][1=\timestep, usedefault]{\glslink{propreachk}{\R^{\prop}_{#1}}}
\newglossaryentry{propreachk}{
	name={\ensuremath{\propreachk}},
	description={Reachable set after propagation step},
	type=reach,
	parent=reachk
}

\newcommandx{\propdrivableareak}[1][1=\timestep, usedefault]{\glslink{propdrivableareak}{\D^{\prop}_{#1}}}
\newglossaryentry{propdrivableareak}{
	name={\ensuremath{\propdrivableareak}},
	description={Drivable area after propagation step},
	type=reach,
	parent=drivableareak
}

\newcommandx{\sBPik}[2][1=(\idxreach), 2=\timestep, usedefault]{\glslink{sBPik}{\baseset^{\prop#1}_{#2}}}
\newglossaryentry{sBPik}{
	name={\ensuremath{\sBPik}},
	description={Propagated base set at time step $\timestep$},
	type=reach,
	parent=sBik
}

\newcommandx{\sAPik}[2][1=(\idxreach), 2=\timestep, usedefault]{\glslink{sAPik}{\aabb^{\prop#1}_{#2}}}
\newglossaryentry{sAPik}{
	name={\ensuremath{\sAPik}},
	description={Projection of $\sBPik$ onto position domain},
	type=reach,
	parent=sAik
}

\newcommandx{\sPPikl}[3][1=(\idxreach), 2=\timestep, usedefault]{\glslink{sPPikllondir}{\polytope^{\prop#1}_{#3, #2}} \glslink{sPPikllatdir}{}}
\newglossaryentry{sPPikllondir}{
	name={\ensuremath{\sPPikl{\curvframelondir}}},
	description={$\idxreach$-th propagated polytope},
	type=reach,
	parent=sPikllondir
}
\newglossaryentry{sPPikllatdir}{
	name={\ensuremath{\sPPikl{\curvframelatdir}}},
	description={$\idxreach$-th propagated polytope},
	type=reach,
	parent=sPikllatdir
}

\newcommand{\repartitioned}{\mathtt{rprt}}
\newcommandx{\repartitioneddrivableareak}[1][1=\timestep, usedefault]{\glslink{repartitioneddrivableareak}{\D^{\repartitioned}_{#1}}}
\newglossaryentry{repartitioneddrivableareak}{
	name={\ensuremath{\repartitioneddrivableareak}},
	description={Drivable area after re-partitioning step},
	type=reach,
	parent=drivableareak,
	sort=10l
}

\newcommandx{\sARik}[2][1=(\idxreach), 2=\timestep, usedefault]{\glslink{sARik}{\aabb^{\repartitioned#1}_{#2}}}
\newglossaryentry{sARik}{
	name={\ensuremath{\sARik[(q)]}},
	description={Part of drivable area after re-partitioning step},
	type=reach,
	parent=sAik,
}

\newcommandx{\sPPhatikl}[3][1=(\idxreach), 2=\timestep, usedefault]{\hat{\polytope}^{#1}_{#3, #2}}

\newcommand{\enlargement}{\glslink{enlargement}{\mathcal{A}^\epsilon}}
\newglossaryentry{enlargement}{
	name={\ensuremath{\enlargement}},
	description={Enlargement to consider shape of ego vehicle},
	type=reach,
	sort=10r
}

\newcommand{\idxconnectedset}{n}
\newcommand{\connected}{\mathcal{C}}

\newcommand{\connectedset}{\glslink{connectedset}{\mathcal{C}}}
\newglossaryentry{connectedset}{
	name={\ensuremath{\connectedset}},
	description={Connected set in the position domain},
	type=drivingcorridor,
	sort=11a
}

\newcommandx{\connectedsetk}[1][1=\timestep, usedefault]{\glslink{connectedsetk}{\connectedset_{#1}}}
\newglossaryentry{connectedsetk}{
	name={\ensuremath{\connectedsetk}},
	description={Connected set at time step $\timestep$},
	type=drivingcorridor,
	parent=connectedset
}

\newcommandx{\connectedsetkn}[2][1=\timestep, 2=\idxconnectedset, usedefault]{\glslink{connectedsetkn}{\connectedset^{(#2)}_{#1}}}
\newglossaryentry{connectedsetkn}{
	name={\ensuremath{\connectedsetkn[][i]}},
	description={$i$-th connected set at time step $\timestep$},
	type=drivingcorridor,
	parent=connectedset
}

\newcommandx{\graphconnectedcomponents}[1][1= , usedefault]{\glslink{graphconnectedcomponents}{\graph_{\connected#1}}}
\newglossaryentry{graphconnectedcomponents}{
	name={\ensuremath{\graphconnectedcomponents}},
	description={Graph storing connected sets},
	type=drivingcorridor,
	sort=11c
}

\newcommand{\node}{\glslink{node}{\mathtt{n}}}
\newglossaryentry{node}{
	name={\ensuremath{\node}},
	description={Node in $\graphconnectedcomponents$},
	type=drivingcorridor,
	parent=graphconnectedcomponents
}

\newcommand{\dc}{C}

\newcommand{\dclon}{\glslink{dclon}{\dc_{\lon}}}
\newglossaryentry{dclon}{
	name={\ensuremath{\dclon}},
	description={Longitudinal driving corridor},
	type=drivingcorridor,
	sort=11e
}

\newcommandx{\dclonk}[1][1=\timestep, usedefault]{\glslink{dclonk}\dc_{\lon, #1}}
\newglossaryentry{dclonk}{
	name={\ensuremath{\dclonk}},
	description={Longitudinal driving corridor at time step $\timestep$},
	type=drivingcorridor,
	parent=dclon
}

\newcommand{\dclat}{\glslink{dclat}{\dc_{\lat}}}
\newglossaryentry{dclat}{
	name={\ensuremath{\dclat}},
	description={Lateral driving corridor},
	type=drivingcorridor,
	sort=11f
}

\newcommandx{\dclatk}[1][1=\timestep, usedefault]{\glslink{dclatk}\dc_{\lat, #1}}
\newglossaryentry{dclatk}{
	name={\ensuremath{\dclatk}},
	description={Lateral driving corridor at time step $\timestep$},
	type=drivingcorridor,
	parent=dclat
}

\newcommandx{\parentset}[1][1=k-1, usedefault]{\mathcal{D}^{\mathrm{parents}}_{#1}}

\newcommand{\distance}{d}
\newcommandx{\latdiscircle}[2][1=\idxcentercircle, 2=\timestep, usedefault]{\glslink{latdiscircle}{\distance^{(#1)}_{#2}}}
\newglossaryentry{latdiscircle}{
	name={\ensuremath{\latdiscircle}},
	description={Lateral distance of $\centercircle$ from $\refpath$ at time step $\timestep$},
	type=constraints,
	sort=12a
}

\newcommandx{\dmink}[2][1=\idxcentercircle, 2=\timestep, usedefault]{\glslink{dmink}{\minval{\distance}^{(#1)}_{#2}}}
\newglossaryentry{dmink}{
	name={\ensuremath{\dmink}},
	description={Minimum admissible value of $\latdiscircle$},
	type=constraints,
	parent=latdiscircle
}

\newcommandx{\dmaxk}[2][1=\idxcentercircle, 2=\timestep, usedefault]{\glslink{dmaxk}{\maxval{\distance}^{(#1)}_{#2}}}
\newglossaryentry{dmaxk}{
	name={\ensuremath{\dmaxk}},
	description={Maximum admissible value of $\latdiscircle$},
	type=constraints,
	parent=latdiscircle
}

\newcommandx{\straightlineik}[3][1=\idxcentercircle, 2=\timestep, usedefault]{\glslink{straightlineik}{g^{(#1)}_{#2}#3}}
\newglossaryentry{straightlineik}{
	name={\ensuremath{\straightlineik{}}},
	description={Straight line perpendicular to $\refpath$ going through $\centercircle$},
	type=constraints,
	sort=12b
}

\newcommandx{\intersectingposik}[2][1=\idxcentercircle, 2=\timestep, usedefault]{\glslink{intersectingposik}{\mathcal{Y}^{(#1)}_{#2}}}
\newglossaryentry{intersectingposik}{
	name={\ensuremath{\intersectingposik}},
	description={Positions in $\dclon$ intersecting with $\straightlineik{}$},
	type=constraints,
	sort=12c
}

\newcommand{\idxinterval}{q}
\newcommandx{\intervaliqk}[3][1=\idxcentercircle, 2=\timestep, 3=\idxinterval, usedefault]{\glslink{intervaliqk}{\mathcal{I}^{(#1)}_{#2,#3}}}
\newglossaryentry{intervaliqk}{
	name={\ensuremath{\intervaliqk}},
	description={$\idxinterval$-th interval of positions in $\dclon$ intersecting $\straightlineik{}$},
	type=constraints,
	sort=12d
}

\newcommandx{\validintervals}[1][1=\idxcentercircle,usedefault]{\mathcal{I}^{(#1)}_{\mathtt{v}}}

\newcommand{\placeholdersys}{\glslink{placeholdersys}{\mathrm{sys}}}
\newglossaryentry{placeholdersys}{
	name={\ensuremath{\placeholdersys}},
	description={Placeholder for a variable},
	sort=a
}

\newcommandx{\proj}[2][1=\lozenge, usedefault]{\glslink{proj}{\mathrm{proj}_{#1}\!(#2 )}}
\newglossaryentry{proj}{
	name={\ensuremath{\proj[\lozenge]{\x}}},
	description={Projection operator that maps the state $\x$ onto its elements $\lozenge$},
	sort=z
}

\newcommandx{\overlap}[1][1=\sAik, usedefault]{\glslink{overlap}{\mathrm{overlap}(#1 )}}
\newglossaryentry{overlap}{
	name={\ensuremath{\overlap}},
	description={Returns all indices $\idxreach$ of the propagated sets $\sBPik$ that overlap with $\sAik$},
	sort=z
}

\newcommand{\convexhull}{\glslink{convexhull}{\mathrm{convexhull}}}
\newglossaryentry{convexhull}{
	name={\ensuremath{\convexhull}},
	description={Computes the convex hull},
	sort=z
}

\newcommand{\signal}{\omega}

\newcommandx{\signalk}[1][1=\timestep, usedefault]{\glslink{signalk}{\signal_{{#1}}}}
\newglossaryentry{signalk}{
	name={\ensuremath{\signalk}},
	description={Signal at time step $\timestep$},
	type=signaltemporallogic,
	sort=14 a
}

\newcommandx{\sminlonARik}[2][1=q, 2=\timestep, usedefault]{\minval{\pos}^{\repartitioned(#1)}_{\curvframelondir, #2}}
\newcommandx{\smaxlonARik}[2][1=q, 2=\timestep, usedefault]{\maxval{\pos}^{\repartitioned(#1)}_{\curvframelondir, #2}}
\newcommandx{\sminlatARik}[2][1=q, 2=\timestep, usedefault]{\minval{\pos}^{\repartitioned(#1)}_{\curvframelatdir, #2}}
\newcommandx{\smaxlatARik}[2][1=q, 2=\timestep, usedefault]{\maxval{\pos}^{\repartitioned(#1)}_{\curvframelatdir, #2}}

\newcommand{\approxsymb}{\mathtt{C}}

\newcommandx{\refcurvaturecirclei}[1][1=i, usedefault]{\curvature_{\refpathsym,#1}^{\approxsymb}}

\setquotestyle{english}

\begin{document}
%
\title{Model Predictive Robustness of Signal Temporal Logic Predicates}
%
%
%


\author{Yuanfei~Lin$^*$,~\IEEEmembership{Student Member,~IEEE,}  Haoxuan Li$^*$, and Matthias Althoff,~\IEEEmembership{Member,~IEEE}%

\thanks{Manuscript received: April 25, 2023; Revised August 04, 2023; Accepted September 22, 2023. This paper was recommended for publication by Associate Editor  T. Asfour and Editor J. Kober upon evaluation of the Associate Editor and Reviewers' comments.
This work was supported by the German Federal Ministry for Digital and Transport (BMDV) within the project \textit{Cooperative Autonomous Driving with Safety Guarantees} (KoSi). } 
\thanks{$^*$ The first two authors have contributed equally to this work. \textit{(Corresponding
		author: Yuanfei Lin.)} The authors are with the School of Computation, Information and Technology, Technical University of Munich, 85748 Garching, Germany (e-mail: {\tt\footnotesize yuanfei.lin@tum.de}; {\tt\footnotesize{haoxuan.li@tum.de}}; {\tt\footnotesize{althoff@tum.de})}.}%
\thanks{Digital Object Identifier 10.1109/LRA.2023.3324582.}
}
%
%

\markboth{IEEE Robotics and Automation Letters. Preprint Version. Accepted October, 2023}
{Lin \MakeLowercase{\textit{et al.}}: Model Predictive Robustness of Signal Temporal Logic Predicates} 

%



\maketitle

\begin{abstract}
The robustness of signal temporal logic not only assesses whether a signal adheres to a specification but also provides a measure of how much a formula is fulfilled or violated. The calculation of robustness is based on evaluating the robustness of underlying predicates. However, the robustness of predicates is usually defined in a model-free way, i.e., without including the system dynamics. Moreover, it is often nontrivial to define the robustness of complicated predicates precisely. To address these issues, we propose a notion of model predictive robustness, which provides a more systematic way of evaluating robustness compared to previous approaches by considering model-based predictions. In particular, we use Gaussian process regression to learn the robustness based on precomputed predictions so that robustness values can be efficiently computed online. We evaluate our approach for the use case of autonomous driving with predicates used in formalized traffic rules on a recorded dataset, which highlights the advantage of our approach compared to traditional approaches in terms of precision. By incorporating our robustness definitions into a trajectory planner, autonomous vehicles obey traffic rules more robustly than human drivers in the dataset. 
\end{abstract}


\section{Introduction}\label{sec:introduction}
\IEEEPARstart{F}{ormal} methods are crucial for specifying and verifying the behavior of autonomous robotic systems \cite{luckcuck2019formal, mehdipour2023formal}. 
Temporal logic, such as linear temporal logic (LTL) \cite{Esterle2020}, metric temporal logic (MTL) \cite{Maierhofer2020a}, and signal temporal logic (STL)~\cite{Arechiga2019}, allows one to specify safety properties and unambiguous tasks for a system over time.\copyrightnotice
 A prominent example is the formalization of traffic rules for autonomous vehicles on which we focus in this paper, e.g., ``\textit{A vehicle must keep a safe distance from others in its lane to avoid collisions in the event of unexpected stops.}" 
MTL and STL are additionally equipped with quantitative semantics, i.e., robustness (aka robustness degree) \cite{FAINEKOS20094262, donze2010robust}, returning the degree of satisfaction or violation of a system with respect to a given specification. In this work, we focus on specifications formalized in STL since one can easily represent discrete-time MTL formulas in STL \cite{bartocci2018specification} and recent research on improving the robustness mainly addresses STL (see Sec.~\ref{subsec:mf-rob}).

\begin{figure}[!t]%
	\centering
	\vspace{2mm}
	\def\svgwidth{\columnwidth}\footnotesize
	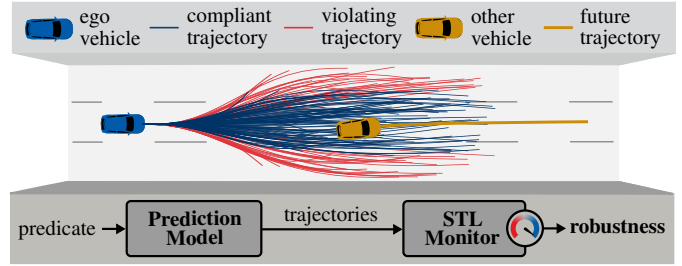
	\caption{Scheme of \RobName robustness computation for the predicate $\mathrm{in\_same\_lane}$. The prediction model generates a finite set of trajectories for all rule-relevant vehicles within a certain time period, of which the rule compliance is checked by an STL monitor. The robustness is calculated based on the future probability of satisfying the predicate.} \label{fig:schema_general}
	\vspace{-2mm}
\end{figure}

The robustness of predicates is an essential building block for evaluating the robustness of STL formulas. However, the robustness of STL predicates 
is typically defined in a model-free manner, i.e., without considering the underlying system dynamics. {Therefore,  when robustness of predicates is inaccurate, the inaccuracy is propagated through the STL formula and thus the overall robustness becomes inaccurate.} 
 In addition, formulating a precise definition of robustness in a unified way can be challenging, especially when not taking into account the unique characteristics and features of the models. 
This issue is addressed in this work by the general idea illustrated in Fig.~\ref{fig:schema_general}. 

%
%
%
%
%

\subsection{Related Work}\label{sec:li_ov}

Our robustness definition aims to facilitate online planning and control with temporal logic specifications. Subsequently, we present related works on specification formalization, model-free robustness, and nonlinear regression approaches.

\subsubsection{Specification Formalization} 
The development of autonomous vehicles requires planning and control to fulfill formal specifications. 
 Several publications formalize traffic rules for interstates \cite{Maierhofer2020a, ruleSTL}, intersections \cite{Maierhofer2022a}, and waterways~\cite{Krasowski2021a} in MTL. 
They use parameterizable Boolean predicates and functions in higher-order logic to specify basic elements of rule specifications. 
In \cite{hekmatnejad2019encoding, Arechiga2019, xiao2021rule}, safety requirements are specified in STL together with the robustness definition for formal verification and controller synthesis. 

\subsubsection{Model-free Robustness} \label{subsec:mf-rob}
The robustness of STL 
is typically nonconvex and nondifferentiable, see, e.g., \cite{donze2010robust}. Therefore, it is generally difficult to deploy fast gradient-based optimization algorithms for online usage, such as optimization-based trajectory planning for autonomous vehicles \cite{YuanfeiLin2022a}. 
Many new extensions for STL robustness have been proposed to address this issue. In \cite{Pant2017, mehdipour2019arithmetic, gilpin2020smooth, Varnai2020updated}, smooth approximations are applied to make the robustness differentiable. 
To handle complex specifications efficiently, \cite{kurtz2022mixed} considers STL formulas over convex predicates.
However, these works only refine the robustness calculation for temporal and logical operators while using affine functions to compute the robustness of predicates. 

For uncertain and changing environments, probabilistic variants of STL are proposed to express safety constraints on random variables and stochastic signals in~\cite{yoo2015control, sadigh2016safe, tiger2020incremental}. 
Similarly, Lee \textit{et. al.}~\cite{lee2021signal} extend STL with uncertain events as predicates to formulate a controller synthesis problem as probabilistic inference, and the authors in \cite{nyberg2021risk} model the risk of violating safety specifications over a random variable.
 


\subsubsection{Nonlinear Regression} 
For safety-critical applications, Gaussian processes (GPs) \cite{williams2006gaussian} have drawn more and more attention to realize accurate predictions 
since they are flexible and powerful for small-data problems. 
 In addition,
GP regression can provide uncertainties for its prediction, which is used to improve the safety and robustness of model predictive control {by online learning} in \cite{kocijan2004gaussian, berkenkamp2015safe, hewing2019cautious}. In this regard, we are inspired by the regression approach described in \cite{torben2022automatic}, where a GP model is used for estimating the robustness of STL specifications including levels of evaluation uncertainty. However, a more general observation space for autonomous driving is adapted for regression in this paper.

\subsection{Contributions}\label{sec:contri}
We present a novel approach to determine the robustness of STL predicates, where the model capability for rule compliance is explicitly considered. When defining a new predicate, the robustness can be directly computed based on its Boolean evaluation instead of relying on manually tuned heuristics. 
In particular, our contributions are:
\begin{enumerate}
	\item proposing a novel systematic robustness measure for STL predicates using predictive models;
	\item using GP regression to learn robustness with comprehensive input features for online applications; and
	\item demonstrating the effectiveness of our robustness definition on formalized traffic rules with real-world data.
\end{enumerate}
The remainder of this paper is organized as follows: In
Sec.~\ref{sec:PRELIMINARIES}, required preliminaries and formulations are introduced. Sec.~\ref{sec:mb_robustness} provides an overview of the model predictive robustness definition and computation. In Sec. \ref{sec:GPR}, GP regression is presented to learn the robustness of predicates. 
We demonstrate the benefits of our method by numerical experiments in Sec.~\ref{sec:case_studies}, followed by conclusions in Sec.~\ref{sec:conc}.

\section{Preliminaries}\label{sec:PRELIMINARIES}
\subsection{System Description and Notation} \label{subsec:veh_config}
\begin{figure}[!t]
	
	\vspace*{2.5mm}
	\def\svgwidth{0.8\columnwidth}\footnotesize	
	\centering
	\mbox{
\begingroup%
  \makeatletter%
  \providecommand\color[2][]{%
    \errmessage{(Inkscape) Color is used for the text in Inkscape, but the package 'color.sty' is not loaded}%
    \renewcommand\color[2][]{}%
  }%
  \providecommand\transparent[1]{%
    \errmessage{(Inkscape) Transparency is used (non-zero) for the text in Inkscape, but the package 'transparent.sty' is not loaded}%
    \renewcommand\transparent[1]{}%
  }%
  \providecommand\rotatebox[2]{#2}%
  \newcommand*\fsize{\dimexpr\f@size pt\relax}%
  \newcommand*\lineheight[1]{\fontsize{\fsize}{#1\fsize}\selectfont}%
  \ifx\svgwidth\undefined%
    \setlength{\unitlength}{280.12923661bp}%
    \ifx\svgscale\undefined%
      \relax%
    \else%
      \setlength{\unitlength}{\unitlength * \real{\svgscale}}%
    \fi%
  \else%
    \setlength{\unitlength}{\svgwidth}%
  \fi%
  \global\let\svgwidth\undefined%
  \global\let\svgscale\undefined%
  \makeatother%
  \begin{picture}(1,0.2759478)%
    \lineheight{1}%
    \setlength\tabcolsep{0pt}%
    \put(0,0){\includegraphics[width=\unitlength,page=1]{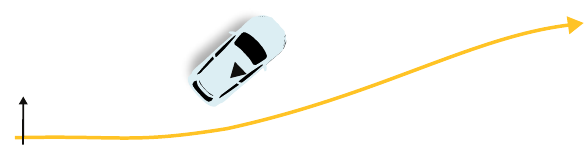}}%
    \put(0.06316831,0.00318254){\color[rgb]{0,0,0}\makebox(0,0)[lt]{\lineheight{1.25}\smash{\begin{tabular}[t]{l}$s$\end{tabular}}}}%
    \put(-0.00179745,0.05977359){\color[rgb]{0,0,0}\makebox(0,0)[lt]{\lineheight{1.25}\smash{\begin{tabular}[t]{l}$d$\end{tabular}}}}%
    \put(0.94016455,0.17497657){\color[rgb]{0,0,0}\makebox(0,0)[lt]{\lineheight{1.25}\smash{\begin{tabular}[t]{l}$\Gamma$\end{tabular}}}}%
    \put(0,0){\includegraphics[width=\unitlength,page=2]{kinematics2.pdf}}%
    \put(0.44130612,0.11146721){\color[rgb]{0,0,0}\makebox(0,0)[lt]{\lineheight{1.25}\smash{\begin{tabular}[t]{l}$d_k$\end{tabular}}}}%
    \put(0.41412157,0.02215695){\color[rgb]{0,0,0}\makebox(0,0)[lt]{\lineheight{1.25}\smash{\begin{tabular}[t]{l}$s_k$\end{tabular}}}}%
    \put(0,0){\includegraphics[width=\unitlength,page=3]{kinematics2.pdf}}%
  \end{picture}%
\endgroup%

	}
	\caption{
		A curvilinear coordinate system aligned with the reference path $\Gamma$.} 
	\vspace{-4mm}
	\label{fig:curvilinear_coordinate_system}
\end{figure}

We model vehicle dynamics as discrete-time systems:
\begin{equation}\label{eq:model}
	x_{k+1} = f(x_{k}, u_{k}),
\end{equation}
where $x_{k}\in\mathbb{R}^{n_x}$ is the state, $u_{k}\in\mathbb{R}^{n_u}$ is the input, 
and the index $k\in\mathbb{N}_0$ is the discrete time step corresponding to the continuous time $t_{k}=k\Delta t$ with a fixed time increment $\Delta t\in\mathbb{R}_{+}$. We use a curvilinear coordinate system \cite{hery2017map} that is aligned with a reference path $\Gamma$ (e.g., the lane centerline), as shown in Fig.~\ref{fig:curvilinear_coordinate_system}. The position of the vehicle at time step $k$ is described by the arc length $s_k$ along $\Gamma$ and the orthogonal deviation $d_k$ from $\Gamma$ at $s_k$. 
The states and the inputs are bounded by sets of permissible values: $\forall k: x_k\in\mathcal{X}_k, u_k\in\mathcal{U}_k$. 
We denote a possible solution of (\ref{eq:model}) at time step $\tau\geq k$ by $\chi(\tau, x_k, u([k, \tau)))$ for an initial state $x_k$ and an input trajectory $u([k, \tau))$. The set of possible state trajectories for the time interval $[k, \tau]$ is denoted as $\boldsymbol{\chi}_{[k, \tau]}$.

We introduce the following sets: $\mathcal{B}\subset\mathbb{N}_0$ is the set of indices referring to rule-relevant obstacles and $\mathcal{L}\subset\mathbb{N}_0$ contains the indices of the occupied lanes by a vehicle. $l^c\in\mathcal{L}$ is the element of $\mathcal{L}$ comprising the lane containing the vehicle center. The road boundary is denoted as $\flat$. Let $\square$ be a variable, we denote its value associated with the ego vehicle, i.e., the vehicle to be controlled, by $\square_{\text{ego}}$ and with other obstacles by $\square_{b}$, with $b\in\mathcal{B}$.
Moreover, the state vector of the ego vehicle and other rule-relevant obstacles at time step $k$ is denoted by $\signal_k\coloneqq[x_{\text{ego},k}, x_{0,k}, \dots, x_{|\mathcal{B}|-1,k}]^{T}\in\mathcal{X}^{|\mathcal{B}|+1}_{k}$.
	In regression models, the inputs and outputs are denoted by ${z}\in\mathbb{R}^{n_z}$ and $y\in\mathbb{R}$, respectively. Given $n_p\in\mathbb{N}_{>0}$ input-output pairs, we define the training set as $\mathcal{D} = \{({z}_i, y_i)\}_{i=1}^{n_p}$. The output predicted by the regression model for the input  $z^*$ is denoted as ${y}^*$.

\subsection{Signal Temporal Logic} \label{subsec:stl}
For traffic rule monitoring, we consider a discrete-time signal $\boldsymbol{\signal} \coloneqq \signal_0\dots\signal_k\dots\signal_{n_{\omega}}\in\boldsymbol{\Omega}$ represented as a sequence of vectors $\signal_k$, where {the set $\boldsymbol{\Omega}$ is a superset that includes all possible signals.} 
Given formulas $\varphi$, $\varphi_1$, and $\varphi_2$, the STL syntax is defined as \cite[Sec. 2.1]{bartocci2018specification}:
\begin{equation}\label{eq:stl}
	\varphi \coloneqq  p \ |\ \lnot \varphi \ |\ \varphi_1 \lor \varphi_2\ |\ \varphi_1 \mathbf{U}_{I} \varphi_2,
\end{equation}
where $p\coloneqq\alpha(\signal_{k})> 0$ is an atomic predicate defined {by the evaluation function $\alpha$$ :\mathcal{X}^{|\mathcal{B}|+1}_{k}$$\rightarrow$$\mathbb{R}$}, 
$\lnot$ and $\lor$ are the Boolean \textit{negation} and \textit{disjunction} operators, respectively, and $\mathbf{U}_I$ is the \textit{until} operator requiring $\varphi_1$ to be true until $\varphi_2$ becomes true in a time interval $I\subseteq\mathbb{N}_0$. 
Other logical connectives and temporal operators can be constructed from (\ref{eq:stl}) such as the \textit{conjunction} operator $\varphi_1 \wedge \varphi_2\coloneqq\lnot(\lnot\varphi_1 \lor \lnot \varphi_2)$ and  the \textit{future} (aka \textit{eventually}) operator $\mathbf{F}_{I}\varphi\coloneqq\top\mathbf{U}_{I}\varphi$ \cite[Sec. 2.1]{bartocci2018specification}.

To show whether an STL formula is satisfied with qualitative semantics, we introduce the characteristic function:
\begin{definition}[Characteristic Function {\cite[Def.~1]{donze2010robust}}]\label{def:char_func}
The characteristic function {$\mathtt{c}:
\boldsymbol{\Omega}\times \mathbb{N}_0\rightarrow\{-1, 1\}$} of an STL formula $\varphi$ (cf.~(\ref{eq:stl})) and a signal $\boldsymbol{\signal}$ at time step $k$ is defined as:
\allowdisplaybreaks
	\begin{align}\label{eq:char_func}
		\mathtt{c}_{p}( \boldsymbol{\signal}, k) &\coloneqq
			\begin{cases}
				1  & \text{if}\  \alpha(\signal_{k})> 0,\\
				-1 & \text{otherwise},
			\end{cases}\nonumber 
		\\ 
		\mathtt{c}_{\lnot\varphi}( \boldsymbol{\signal}, k) &\coloneqq -\mathtt{c}_{\varphi}( \boldsymbol{\signal}, k),\nonumber \\
		{\mathtt{c}_{\varphi_1\lor\varphi_2}(\boldsymbol{\signal}, k)} &\coloneqq \max\big(\mathtt{c}_{\varphi_1}(\boldsymbol{\signal}, k),\mathtt{c}_{\varphi_2}(\boldsymbol{\signal}, k)\big),\\
		\mathtt{c}_{\varphi_1\mathbf{U}_I\varphi_2}(\boldsymbol{\signal}, k) &\coloneqq \max_{\tau\in (k+I)\cap\mathbb{N}_0}\big(\min\big(\mathtt{c}_{\varphi_2}(\boldsymbol{\signal},\tau),\nonumber \\ 
		& \quad\qquad\quad\quad\hspace{5mm} \qquad\min_{\tau'\in[k,\tau)}\!\mathtt{c}_{\varphi_1}(\boldsymbol{\signal},\tau')\big)\big).\nonumber 
	\end{align}

\end{definition}
For an exemplary calculation of the model-free STL robustness, we refer interested readers to \cite[Def.~3]{donze2010robust}.

\subsection{Problem Formulation}\label{subsec:prob_form}
Inspired by the metrics described in \cite{mehdipour2019arithmetic, nyberg2021risk}, the robustness of STL predicates should follow the subsequent properties for its application to planning and control problems:
\begin{property}[Soundness]\label{prop:soundness}
	{The robustness is sound when positive robustness ensures necessary and sufficient predicate satisfaction, and likewise, negative robustness identifies predicate violations.}
\end{property}
\begin{property}[Smoothness]\label{prop:smoothness}
	The robustness is smooth with respect to its input almost everywhere\footnote{This is because smoothness across the entire
		domain is a too strict requirement  for STL robustness  \cite[Sec.~\Romannum{4}]{Varnai2020updated}.}, except on the satisfaction or violation boundaries {where $\alpha(\signal_{k})=0$}. 
\end{property}
\begin{property}[Monotonicity]\label{prop:Monotonicity}
	The robustness is monotonic, i.e., increasing with higher probabilities of satisfying the predicate and decreasing otherwise.
\end{property}
{For the mathematical formalization of these properties, we refer the reader to \cite[Prop. 1 and Prop. 3]{Varnai2020updated} and \cite[Sec.~3.1]{majumdar2020should}.} The soundness and smoothness can be considered by following the requirements in \cite[Thm.~1 and Prop.~1]{mehdipour2019arithmetic}. For monotonicity, the robustness needs to rely on system dynamic models and predictive behaviors, which we call \textit{\RobName robustness} since the idea is inspired by model predictive control~\cite{camacho2013model} and {model-based criticality measures~\cite{schneider2021towards}}. {In this work, our aim is to define robustness that satisfies all the desired properties for STL predicates, primarily focusing on traffic rule predicates defined in higher-order logic. However, our definition can be extended to other predicate types.}

\section{Model Predictive Robustness}\label{sec:mb_robustness}
In this section, model predictive robustness is first formally defined. 
Then we introduce a Bayesian representation of the predictive model. Afterwards, the overall algorithm for the computation of the robustness is presented, followed by its detailed description. 

\subsection{Definition}\label{subsec: definition}
{At time step $\tau\geq k$, we denote the predicted signal vector and the output of the characteristic function as $\omega'_\tau$ and $C^p_{\tau}$, respectively, which are modeled as
random variables as their future evaluation is uncertain.}
\begin{definition}[Model Predictive Robustness]
	The \RobName robustness $\rho^{\MP}_p$ considers the probability ${P}(\boldsymbol{\omega}, k)$ that the output of the characteristic function is unchanged, {i.e., $C^{p}_{\tau}=\!\mathtt{c}_{p}( \boldsymbol{\signal}, k)$} over a finite prediction horizon $h\in\mathbb{N}_0$ and is defined as: 
	\allowdisplaybreaks
\begin{align}\label{eq:mb-robustness}
			\rho^{\MP}_{p}( \boldsymbol{\signal}, k) \coloneqq& 
			\begin{dcases}
				 \frac{{P}(\boldsymbol{\signal}, k) - \bar{P}_{+,\min}}{\bar{P}_{+,\max}-\bar{P}_{+,\min}} & \text{if}\  {\mathtt{c}_{p}( \boldsymbol{\signal}, k)} = 1,\\
				-\frac{{P}(\boldsymbol{\signal}, k) - \bar{P}_{-,\min}}{\bar{P}_{-,\max}-\bar{P}_{-,\min}} & \text{if}\  {\mathtt{c}_{p}( \boldsymbol{\signal}, k)} = -1,\\
			\end{dcases}
			\nonumber\\
			\text{s.t.}\ {P}(\boldsymbol{\signal},k) \coloneqq&\frac{1}{h+1}\sum_{\tau=k}^{k+h}P\big(C^p_{\tau}=\mathtt{c}_{p}( \boldsymbol{\signal}, k)\big),\\ 
			\bar{P}_{+,\max/\min}\! \coloneqq&\underset{\bar{\boldsymbol{\omega}}\in \boldsymbol{\Omega},\bar{k}\in\mathbb{N}_0}{\max/\min}\ {P}(\bar{\boldsymbol{\signal}},\bar{k} \ |\ \mathtt{c}_{p}( \bar{\boldsymbol{\signal}}, \bar{k})=1),\nonumber\\
			\bar{P}_{-,\max/\min}\! \coloneqq&\underset{\bar{\boldsymbol{\omega}}\in \boldsymbol{\Omega},\bar{k}\in\mathbb{N}_0}{\max/\min}\ {P}(\bar{\boldsymbol{\signal}},\bar{k} \ |\ \mathtt{c}_{p}( \bar{\boldsymbol{\signal}}, \bar{k})=-1),\nonumber
		\end{align}
	where 	$\rho^{\MP}_{p}( \boldsymbol{\signal}, k)$ is normalized to the interval $[-1, 1]$ with the normalization values $\bar{P}_{+,\max/\min}\in\mathbb{R}_0$ and $\bar{P}_{-,\max/\min}\in\mathbb{R}_0$, { $ \bar{\boldsymbol{\signal}}$ and $\bar{k}$ are a possible signal and time step, respectively, and ${P}(\bar{\boldsymbol{\signal}},\bar{k} \ |\ \mathtt{c}_{p}( \bar{\boldsymbol{\signal}}, \bar{k}))$ is the conditional probability of ${P}(\bar{\boldsymbol{\signal}},\bar{k})$ given the value of $\mathtt{c}_{p}( \bar{\boldsymbol{\signal}}, \bar{k})$}. {Note that the normalization values can be approximated for online use,  
	e.g., from a dataset.} 
	
\end{definition}


\subsection{Prediction Model with Monte Carlo Simulation}\label{subsec:mc}
The characteristic functions within the prediction horizon, as described in (\ref{eq:mb-robustness}), can be computed by performing inference on hidden variables \cite[Pt. I]{koller2009probabilistic}. Therefore, the probability of maintaining the value of the characteristic function at time step $\tau$ can be written as:
	   \begin{equation}\label{eq:probability_rewrite2}
	\begin{split}
		\!\!\!{P}\big(C^{p}_{\tau}
			=\mathtt{c}_{p}( \boldsymbol{\signal}, k)\big)
			\!&=\!\!\sum_{{\signal}'_{\tau}\in\Omega'_{\tau}}\!\!\!{P}\big(C^{p}_{\tau}=\mathtt{c}_{p}( \boldsymbol{\signal}, k)\  |\ {\signal}'_{\tau}\big)P({\signal}'_{\tau})
			\\\! & =\!\! \sum_{{\signal}'_{\tau}\in\Omega'^{\mathtt{c}}_{\tau}}\!\!\!P({\signal}'_{\tau}),
		\end{split}
	\end{equation}
{where $P({\signal}'_{\tau})$ is the probability of ${\signal}'_{\tau}$, ${P}(C^{p}_{\tau}\!=\!\mathtt{c}_{p}( \boldsymbol{\signal}, k) \ |\ \signal'_{\tau})$ is the observation probability with values of either $0$ or $1$, $\Omega'_{\tau}$ is the set of all $\signal_{\tau}'$, and $\Omega'^{\mathtt{c}}_{\tau}\subseteq\Omega'_{\tau}$ is its subset satisfying $C^{p}_{\tau}=\!\mathtt{c}_{p}( \boldsymbol{\signal}, k)$.} 
However, it is nontrivial to obtain the exact distribution of $P({\signal}'_{\tau})$ in (\ref{eq:probability_rewrite2}) \cite[Sec.~V]{althoff2009model}. 
Instead, we employ Monte Carlo simulation to generate potential future behaviors of the ego vehicle \cite[Sec.~1]{spanos2019probability}, while using predicted {or recorded} behaviors for other traffic participants 
\cite{broadhurst2005monte}.
Note that such a choice is not mandatory and any other estimation method that approximates the true distribution of $P({\signal}'_{\tau})$  suffices. 
By assembling the obtained trajectories, 
 the 
 Monte Carlo 
estimation of the probability in (\ref{eq:probability_rewrite2}) 
 can be calculated as the ratio of the number of predicted signal vectors in $\Omega'^{\mathtt{c}}_{\tau}$ to the size of $\Omega'_{\tau}$ 
 \cite[Sec.~1]{spanos2019probability}:
\begin{equation}\label{eq:rob_tau}
	\sum_{{\signal}'_{\tau}\in\Omega'^{\mathtt{c}}_{\tau}}P({\signal}'_{\tau})\approx\frac{|\Omega'^{\mathtt{c}}_{\tau}|}{|{\Omega'_{\tau}}|}.
\end{equation} 

\subsection{Overall Algorithm}\label{subsec:overall}
Alg.~\ref{alg:main} provides an overview of the computation of the \RobName robustness for traffic rule predicates. At time step $k$, we receive as input the predicate, the signal vector, and the prediction horizon. 
First, we randomly sample a set of trajectories $\boldsymbol{\chi}_{\text{ego},{[k, k + h]}}$ for the ego vehicle based on its current state (line~\ref{alg1:ego}; cf. Sec.~\ref{subsec:sampling_based}). Afterward, if the predicate depends also on other traffic participants, i.e.,  $|\mathcal{B}|>0$, their future trajectories $\boldsymbol{\chi}_{\mathcal{B},{[k, k + h]}}$ are obtained (line~\ref{alg1:other}). 
When computing the
robustness offline, we use their recorded trajectories. 
Then the set of all predicted signals ${\Omega}'_{[k,k+h]}\coloneqq\{\Omega'_{k}\dots\Omega'_{\tau}\dots\Omega'_{k+h}\}$ is constructed by {all possible combinations of elements in $\boldsymbol{\chi}_{\text{ego},{[k, k + h]}}$ and $\boldsymbol{\chi}_{\mathcal{B},{[k, k + h]}}$} (line~\ref{alg1:product}).
The last step is to compute the model predictive robustness by checking the relative frequency of rule-compliant or rule-violating predicted signals with an STL monitor (line~\ref{alg1:compute}).
\renewcommand{\algorithmicrequire}{\textbf{Input:}}
\renewcommand{\algorithmicensure}{\textbf{Output:}}

\begin{algorithm}[h!]
	\small
		\begin{algorithmic}[1]
			\Require predicate $p$, signal $\boldsymbol{\signal}$ at time step $k$, horizon $h$
			\Ensure \RobName robustness $\rho^{\MP}_p$\vspace{.mm}
			\State $\boldsymbol{\chi}_{\text{ego}, {[k, k + h]}}$ $\gets$ \Call{sampleEgoTraj}{$\signal_{k}$, $h$} \Comment{Sec.~\ref{subsec:sampling_based}} \label{alg1:ego} \vspace{.5mm}
			\If{$|\mathcal{B}|>0$}
			\State $\boldsymbol{\chi}_{\mathcal{B}, {[k, k + h]}}$ $\gets$ \Call{{obtainOtherTraj}}{$\signal_{k}$, $h$} \label{alg1:other}\vspace{.5mm}
			\Else\vspace{.5mm}
			\State $\boldsymbol{\chi}_{\mathcal{B}, {[k, k + h]}}$ $\gets$ $\emptyset$\vspace{.5mm}
			\EndIf\vspace{.5mm}
			\State ${\Omega}'_{[k,k+h]}$ $\gets$ \Call{constructSignal}{$\boldsymbol{\chi}_{\text{ego}, {[k, k + h]}}, \boldsymbol{\chi}_{\mathcal{B},{[k, k + h]}}$} \label{alg1:product}
			\vspace{.5mm}
			\State  $\rho^{\MP}_p$ $\gets$ \Call{stlMonitoring}{$p$, ${\Omega}'_{[k,k+h]}$} \label{alg1:compute} \Comment{(\ref{eq:mb-robustness}), (\ref{eq:rob_tau})}\vspace{.5mm}
			\State \Return $\rho^{\MP}_p$ \vspace{-.5mm}
	\end{algorithmic}
	\caption{\small\textsc{computeModelPredictiveRobustness}}
	\label{alg:main}
\end{algorithm}
\subsection{Trajectory Sampling for the Ego Vehicle}\label{subsec:sampling_based}
\begin{figure}[h!]	
	\captionsetup[subfigure]{aboveskip=+1mm,belowskip=+1mm}
	\centering
	\vspace{-2.5mm}
	\begin{subfigure}[b]{0.49\linewidth}
		\footnotesize
		\centering
		\def\svgwidth{0.95\columnwidth}
		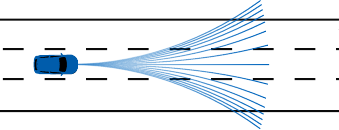
		\caption{{\footnotesize Input space.}}\label{fig:ibs}
	\end{subfigure}
	\begin{subfigure}[b]{0.49\linewidth}
		\footnotesize
		\centering
		\def\svgwidth{0.95\columnwidth}
\begingroup%
  \makeatletter%
  \providecommand\color[2][]{%
    \errmessage{(Inkscape) Color is used for the text in Inkscape, but the package 'color.sty' is not loaded}%
    \renewcommand\color[2][]{}%
  }%
  \providecommand\transparent[1]{%
    \errmessage{(Inkscape) Transparency is used (non-zero) for the text in Inkscape, but the package 'transparent.sty' is not loaded}%
    \renewcommand\transparent[1]{}%
  }%
  \providecommand\rotatebox[2]{#2}%
  \newcommand*\fsize{\dimexpr\f@size pt\relax}%
  \newcommand*\lineheight[1]{\fontsize{\fsize}{#1\fsize}\selectfont}%
  \ifx\svgwidth\undefined%
    \setlength{\unitlength}{162.35410197bp}%
    \ifx\svgscale\undefined%
      \relax%
    \else%
      \setlength{\unitlength}{\unitlength * \real{\svgscale}}%
    \fi%
  \else%
    \setlength{\unitlength}{\svgwidth}%
  \fi%
  \global\let\svgwidth\undefined%
  \global\let\svgscale\undefined%
  \makeatother%
  \begin{picture}(1,0.27638288)%
    \lineheight{1}%
    \setlength\tabcolsep{0pt}%
    \put(0,0){\includegraphics[width=\unitlength,page=1]{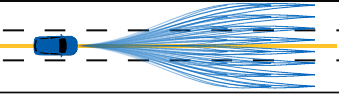}}%
    \put(0.01683622,0.01456713){\color[rgb]{0,0,0}\makebox(0,0)[lt]{\lineheight{1.25}\smash{\begin{tabular}[t]{l}$\Gamma$\end{tabular}}}}%
    \put(0,0){\includegraphics[width=\unitlength,page=2]{SBS.pdf}}%
    \put(0.25265932,0.02890176){\color[rgb]{0,0,0}\makebox(0,0)[lt]{\lineheight{1.25}\smash{\begin{tabular}[t]{l}end state\end{tabular}}}}%
  \end{picture}%
\endgroup%
\vspace{2.15mm}
		\caption{{\footnotesize State space.}}\label{fig:sbs}
	\end{subfigure}
	\vspace{-6.5mm}
	\caption{Comparison of different sampling strategies for the ego vehicle.}\label{fig:sampling}
	\vspace{-4mm}
\end{figure}

Trajectories can be sampled either in input space or in state space \cite{howard2008state} (see Fig.~\ref{fig:sampling}).
The former generates trajectories through the forward simulation of the vehicle dynamics. 
In contrast, with the state-based strategy, the trajectories are obtained by connecting pairs of vehicle states, which helps to exploit environmental constraints to avoid unnecessary samples. 
Since we only consider structured environments and want to provide more reactive capabilities for the ego vehicle, the state-space sampling approach described in \cite{werling2012optimal} is used. 

As shown in Fig.~\ref{fig:sbs}, {a set of end states at time step $k+h$ are drawn uniformly from predefined sampling intervals} in the curvilinear coordinate system along the reference path $\Gamma$. {Alternative sampling distributions can be adapted if a specific pattern of the ego vehicle's behavior is known.} 
 Then trajectories are computed by connecting the end states with the current ego state at time step $k$ using quintic polynomials. 
Afterward, the kinematic feasibility of the trajectories is checked, e.g., using the \textit{CommonRoad Drivability Checker} \cite{PekIV20}, and the feasible ones are transformed from the curvilinear coordinate to the Cartesian frame as sampled trajectories for the ego vehicle. 

\section{Gaussian Process Regression}\label{sec:GPR}
Determining the \RobName robustness is computationally expensive as well as noisy {due to the sampling errors (cf. Sec.~\ref{subsec:sampling_based}).} 
{Moreover, the future trajectories of vehicles are usually not directly accessible during online usage.
As motivated in Sec.~\ref{sec:li_ov}, we choose GP regression to learn the robustness, addressing the needs of online applications.}
\subsection{Feature Variables and Robustness Prediction}\label{subsec:fea_var}
A GP is a collection of random variables such that any finite subset of those variables is jointly Gaussian distributed~\cite[Def.~2.1]{williams2006gaussian}. 
The discriminative capabilities of GP regression models highly depend on the selection of the feature variables~\cite[Sec.~7.5]{williams2006gaussian}. 
We group the measurements of the feature variables ${z}$ in four categories as listed in Tab.~\ref{tab:feature_variables}, which are either rule-related or commonly used for learning-based algorithms, e.g., in \cite{commonroad-rl, gindele2010probabilistic}. 
{It should be noted that the feature variables pertain solely to the evaluating time step. Consequently, there is no necessity for Monte Carlo simulation, as discussed in Sec.~\ref{subsec:mc}, for inferring robustness online.}
To avoid false negatives and false positives of the robustness prediction, i.e., to ensure the soundness  (cf. Prop.~\ref{prop:soundness}), we rectify the regression output 
using the characteristic function (cf. Def.~\ref{def:char_func}) to obtain the estimated \RobName robustness $\tilde{\rho}^{\MP}_p$ 
as:
\begin{equation}\label{eq:clip}
	\tilde{\rho}^{\MP}_{p}( \boldsymbol{\signal}, k)=
	\begin{dcases}
		\widetilde{\max}({y}^*
		, 0) & \text{if}\  {\mathtt{c}_{p}( \boldsymbol{\signal}, k)} = 1,\\
		\widetilde{\min}({y}^*
		, 0) & 	\text{if}\  {\mathtt{c}_{p}(\boldsymbol{\signal}, k)} = -1,\\
	\end{dcases} 
\end{equation}
where the value of the predicted robustness with incorrect signs is set to $0$ {using the smooth minimum and maximum operators $\widetilde{\min}$ and $\widetilde{\max}$ defined in \cite[(9) and (11)]{gilpin2020smooth}.} 

\renewcommand{\arraystretch}{1.12}
\begin{table}[h!]\small
	\begin{center}
		\vspace{0mm}
		\caption{Feature variable definitions. All values presented are in SI units and at time step $k$ unless otherwise specified. To compute $\Delta_{\flat}$ or $\Delta_{{l}^c}$, we use the signed distance from the vehicle center to its closest point at the road boundary $\flat$ or the bounds of 
			lane ${l}^c$.}
		\label{tab:feature_variables}
		\begin{tabular}{p{2.3cm}p{5.5cm}}
			\toprule[1.pt]
			\textbf{Feature Variable}  & \textbf{Description} \\
			\midrule
			\multicolumn{2}{l}{\small \textbf{Rule-Related}} \\
			
			${\mathtt{c}}_{p}( \boldsymbol{\signal}, k)$ & characteristic function\\
			\midrule
			\multicolumn{2}{l}{\textbf{Ego-Vehicle-Related}} \\
			
			$\ell_{\text{ego}}$, $w_{\text{ego}}$	     & vehicle length and width\\
			$x_{\text{ego}}$, $u_{\text{ego}}$ & state and input\\
			$\Delta_{{l}_{\mathtt{l}}^c, \text{ego}}$, $\Delta_{{l}_{\mathtt{r}}^c, \text{ego}}$            & left and right distance to the left and right boundary of the occupied lane\\
			$\Delta_{\flat_{\mathtt{l}}, \text{ego}}$, $\Delta_{\flat_{\mathtt{r}}, \text{ego}}$ & distance to the left and right road boundary\\
			\midrule
			\multicolumn{2}{l}{\textbf{Other-Vehicle-Related}} \\
			
			$\ell_{b}$, $w_{b}$				         & vehicle length and width\\
			$x_{b}$ & state\\
			$\Delta_{{l}_{{l}}^c, b}$, $\Delta_{{l}_{\mathtt{r}}^c, b}$            & left and right distance to the left and right boundary of the occupied lane\\
			$\Delta_{\flat_{\mathtt{l}}, b}$, $\Delta_{\flat_{\mathtt{r}}, b}$ & distance to the left and right road boundary\\
			\midrule
			\multicolumn{2}{l}{\textbf{Ego-Other-Relative}} \\
			
			$\Delta s$, $\Delta d$ & relative longitudinal and lateral distance \\
			$\Delta v$ & relative velocity\\
			\bottomrule[1.pt]
		\end{tabular}
		\vspace{-5mm}
	\end{center}
\end{table}
\subsection{Evaluation on Properties}

\renewcommand{\arraystretch}{1.1}
\newcolumntype{P}[1]{>{\centering\arraybackslash}p{#1}}
\begin{table}[t!]
	\begin{center}\small
		\vspace{2.5mm}
		\caption{Comparison of fulfilled properties. Note that we examine the satisfaction of properties for all predicates.}
		\label{tab:predicates}
		\begin{tabular}{p{1.8cm}P{2.75cm}P{2.75cm}}
			\toprule[1.1pt]
			\textbf{Property}  & {\centering \textbf{Model-free \cite{ruleSTL}}}  & \textbf{Model Predictive}\\			\midrule
			\textbf{Soundness} & \textcolor{TUMblue}{\cmark} & \textcolor{TUMblue}{\cmark} \\
			\textbf{Smoothness} & \textcolor{RPTHred}{\xmark} & \textcolor{TUMblue}{\cmark} \\
			\textbf{Monotonicity} & \textcolor{RPTHred}{\xmark} & \textcolor{TUMblue}{\cmark} \\
			\bottomrule[1.1pt] 
		\end{tabular}
	\end{center}
\vspace{-4mm}
\end{table}
Tab.~\ref{tab:predicates} summarizes the properties of the model-free and \RobName robustness.  {The model-free robustness is denoted as ${\rho}^{\mathtt{MF}}$ and  defined in \cite{ruleSTL}.} It shows that all desired properties listed in Sec.~\ref{subsec:prob_form} 
are fulfilled by the model predictive robustness.

\begin{theorem} \label{theorem:property}
	{The model predictive robustness, as defined by (\ref{eq:mb-robustness}), inherently satisfies Props. \ref{prop:soundness} and \ref{prop:Monotonicity} and substantiates Prop. \ref{prop:smoothness} when combined with GP regression.}
\end{theorem}
\begin{IEEEproof}
	The monotonicity and soundness of the model predictive robustness hold by definition. The smoothness of the robustness is fulfilled using the GP regression. A GP is fully specified by its mean and covariance (aka kernel) function. The mean is typically assumed to be zero in practice 
	and we choose a squared-exponential kernel function, which is built on the assumption that feature variables close to each other (in terms of squared Euclidean distance) have similar robustnesses. The covariance between any two input values ${z}_i$ and ${{z}}_{i'}$, $i, i'\in\{1,\dots,n_p\}$, is then given by \cite[(6)]{berkenkamp2015safe}:
	\begin{equation}\label{eq:se-kernel}
		\mathtt{k}({z}_i, {{z}}_{i'})\!=\!\sigma_\rho^2\exp\big(\frac{1}{2}({z}_i-{{z}}_{i'})^{T}\!L_{\rho}^{-1}({z}_i-{{z}}_{i'})\big)+\delta(i, {i'})\sigma_{\delta}^2,
	\end{equation}
where $L_{\rho}$ is the diagonal length-scale matrix with positive values, $\delta(\cdot)$ is the Kronecker delta function, and $\sigma_\rho$ and $\sigma_\delta$ are the process deviation and discretization noise, respectively. 
The prediction of the regression model
is a conditional probability distribution of $y^*$ given $\mathcal{D}$, which remains a
Gaussian distribution with ${P}(y^*|\mathcal{D})=\mathcal{N}(\mu({z}^*),\sigma^2({z}^*))$ \cite[Sec. 2.2]{williams2006gaussian} and 
\begin{equation}\label{eq:mean_cov}
	\begin{split}
		\mu({z}^*) &= \mathbf{k}^T({z}^*)\mathtt{K}^{-1}\mathbf{y},\\
		\sigma^2({z}^*) &= \mathtt{k}({z}^*, {z}^*)-\mathbf{k}^T({z}^*)\mathtt{K}^{-1} \mathbf{k}({z}^*),
	\end{split}
\end{equation}
where $\mathbf{y}=[y_1,\dots,y_{n_p}]^{T}$ is the vector of observed outputs, $\mathtt{K}\in\mathbb{R}^{n_p\times n_p}$ is the covariance matrix with entries $\mathtt{K}_{i,i'}=\mathtt{k}({z}_i, {z}_{i'})$, and $\mathbf{k}({z}^*)=[\mathtt{k}({z}_1, {z}^*),\dots,\mathtt{k}({z}_{n_p}, {z}^*)]^T$ contains the covariances evaluated at all training data and observation pairs. We take the mean $\mu({z}^*)$ as the computed model predictive robustness, i.e., $y^*=\mu({z}^*)$. Based on (\ref{eq:se-kernel}) and (\ref{eq:mean_cov}), $y^*$ is
a linear combination of the squared-exponential kernel functions contained in $\mathbf{k}({z}^*)$, which are infinitely differentiable with respect to ${z}^*$ \cite[Sec. 4.2.1]{williams2006gaussian}. As a result, the model predictive robustness is smooth. Note that even if the rectification in (\ref{eq:clip}) is used, the computed robustness is still sound and smooth (see \cite[Thm. 1]{gilpin2020smooth}). 
\end{IEEEproof}

With the deviation $\sigma({z}^*)$ in (\ref{eq:mean_cov}), one can obtain the confidence intervals of the robustness computation. Since $\mathtt{K}$ only needs to be inverted once for a given dataset, the complexities of evaluating the mean and variance in (\ref{eq:mean_cov}) are both $\mathcal{O}(n_p^2)$ \cite[Alg.~2.1]{williams2006gaussian}.
For applications with large amounts of data points, a sparse approximation of the GP regression \cite{quinonero2005unifying} can be used to reduce the computational complexity. 

\section{Numerical Experiments}\label{sec:case_studies}

%

We evaluate the applicability and efficiency of the \RobName robustness using the highD dataset \cite{highd} 
and German interstate traffic rules from \cite{Maierhofer2020a, ruleSTL}. 
Our simulation is based on \textit{CommonRoad} \cite{Althoff2017a} and \textit{GPyTorch} \cite{gardner2018gpytorch} is used to model and solve the GP regression. 
We use the vehicle model from \cite[(8) and (13)]{pek2020fail}, which separates the longitudinal and lateral motion of the vehicle in the curvilinear coordinate system (cf. Fig.~\ref{fig:curvilinear_coordinate_system}).
The trajectories for all rule-relevant vehicles described in Sec.~\ref{sec:mb_robustness} are determined with a time horizon of $h\Delta t = 1.5 s$ and the step size $\Delta t = 0.04 s$. {The number of sampled trajectories for the ego vehicle is set to $1,000$, chosen after experimenting with different sample sizes to ensure a relatively small variance in the computed robustness, considering the computational complexity involved}\footnote{{We independently sample the velocity, the lateral position, and the derivative of lateral position of the end states within the intervals $[v_k-17.25, v_k+17.25]$ in $m/s$, $[d_k-5, d_k+5]$ in $m$, and $[-3, 3]$ in $m/s$, respectively.}}. 
The code used in this paper is published as open-source at  \url{https://gitlab.lrz.de/tum-cps/mpr}. {Our experiment videos can be accessed in the supplementary files of the paper.}

\subsection{GP Regression Model}
{Our robustness definition is validated by its usefulness against all predicates in the German interstate rules as referenced in \cite{ruleSTL}. We obtain $12,500$ and $3,750$ data points from the highD scenarios for predicates with $|\mathcal{B}|>0$ and $|\mathcal{B}|=0$, respectively. The data points are randomly divided into two sets, the training set and the test set, in a $4:1$ ratio. 
		\begin{table}[t!]
		\renewcommand{\arraystretch}{1.12}
		\begin{center}\footnotesize
			\vspace{2.5mm}
			\caption{{Parameters and evaluation results for considered predicates {which are formally defined in \cite{Maierhofer2020a}}. If there are no data points in the dataset that satisfy or violate the predicate, we set the normalization values $\bar{P}_{+,\min}$ and $\bar{P}_{+,\max}$ to $0$ and $1$, respectively, and similarly for $\bar{P}_{-,\min}$ and $\bar{P}_{-,\max}$. {The measures represent the distribution of model-free and model predictive robustness among $1,000$ randomly selected data points, where the span denotes the difference between the maximum and minimum values. The larger values for variance and span are marked with bold numbers.}}}	\label{tab:appendix}
			\begin{tabular}{@{}lc@{\hspace{5pt}}cc@{\hspace{3pt}}cc@{\hspace{5pt}}c@{}}
				\toprule[1.0pt]
				\textbf{Data}  & \multicolumn{2}{c}{$\boldsymbol{\mathrm{in\_same\_lane}}$} & \multicolumn{2}{c}{$\boldsymbol{\mathrm{single\_lane}}$} & \multicolumn{2}{c}{$\boldsymbol{\mathrm{in\_front\_of}}$} \\\hline
				 \noalign{\vspace{0.2ex}} ${\bar{P}_{+,\min}}$ & \multicolumn{2}{c}{$0.1935$} & \multicolumn{2}{c}{$0.2326$} & \multicolumn{2}{c}{$0.0526$}\\
				${\bar{P}_{+,\max}}$ & \multicolumn{2}{c}{$0.9933$} & \multicolumn{2}{c}{$0.8076$} & \multicolumn{2}{c}{$1.0000$}\\ ${\bar{P}_{-,\min}}$ & \multicolumn{2}{c}{$0.0000$} & \multicolumn{2}{c}{$0.0763$} & \multicolumn{2}{c}{$0.0000$}\\
				${\bar{P}_{-,\max}}$ & \multicolumn{2}{c}{$0.9286$} & \multicolumn{2}{c}{$0.6072$} & \multicolumn{2}{c}{$0.9737$}\\
				Comp. Time & 
				\multicolumn{2}{c}{$4.07 ms$} & \multicolumn{2}{c}{$3.99 ms$} & \multicolumn{2}{c}{$4.12 ms$}\\		
				Precision & \multicolumn{2}{c}{$0.9991$} & \multicolumn{2}{c}{$1.0000$} &  \multicolumn{2}{c}{$1.0000$}\\
				Recall & \multicolumn{2}{c}{$1.0000$} &  \multicolumn{2}{c}{$1.0000$}&  \multicolumn{2}{c}{$0.9992$}\\
				F1-score & \multicolumn{2}{c}{$0.9995$} &  \multicolumn{2}{c}{$1.0000$} &  \multicolumn{2}{c}{$0.9996$}\\\hline 
				{\textbf{Measure}}& $\rho^{\mathtt{MF}}$ & $\rho^{\mathtt{MP}}$	& $\rho^{\mathtt{MF}}$ & $\rho^{\mathtt{MP}}$	& $\rho^{\mathtt{MF}}$ & $\rho^{\mathtt{MP}}$\\\hline
				Variance & $0.0166$ & $\boldsymbol{0.6017}$ & $0.0005$ & $\boldsymbol{0.0798}$ & $0.4523$ & $\boldsymbol{0.9868}$\\
				Span & $0.4713$ & $\boldsymbol{1.9628}$ & $0.1787$ & $\boldsymbol{1.9764}$ & $2.0000$ & $\boldsymbol{2.0000}$ \\
				\midrule	
				\textbf{Data}  & \multicolumn{2}{c}{$\boldsymbol{\mathrm{cut\_in}}$} & \multicolumn{2}{c}{$\makecell[c]{\boldsymbol{\mathrm{keeps\_safe\_}}\\\boldsymbol{\mathrm{distance\_prec}}}$} & \multicolumn{2}{c}{$\makecell[c]{\boldsymbol{\mathrm{brakes\_}}\\\boldsymbol{\mathrm{abruptly}}}$} \\\hline 
				\noalign{\vspace{0.2ex}}${\bar{P}_{+,\min}}$ & \multicolumn{2}{c}{$0.0650$} & \multicolumn{2}{c}{$0.1959$} & \multicolumn{2}{c}{$0.3870$}\\
				${\bar{P}_{+,\max}}$ & \multicolumn{2}{c}{$0.4575$} & \multicolumn{2}{c}{$1.0000$} & \multicolumn{2}{c}{$0.4154$}\\ 
				${\bar{P}_{-,\min}}$ & \multicolumn{2}{c}{$0.0000$} & \multicolumn{2}{c}{$0.0000$} & \multicolumn{2}{c}{$0.2619$}\\
				${\bar{P}_{-,\max}}$ & \multicolumn{2}{c}{$0.4347$} & \multicolumn{2}{c}{$0.7796$} & \multicolumn{2}{c}{$0.5794$}\\
				Comp. Time & \multicolumn{2}{c}{$4.03 ms$} & \multicolumn{2}{c}{$4.13 ms$} & \multicolumn{2}{c}{$3.97 ms$}\\		
				Precision & \multicolumn{2}{c}{$1.0000$} &  \multicolumn{2}{c}{$1.0000$} & \multicolumn{2}{c}{$1.0000$}\\
				Recall & \multicolumn{2}{c}{$1.0000$} &  \multicolumn{2}{c}{$1.0000$} & \multicolumn{2}{c}{$1.0000$}\\
				F1-score & \multicolumn{2}{c}{$1.0000$} &  \multicolumn{2}{c}{$1.0000$} & \multicolumn{2}{c}{$1.0000$} \\\hline {\textbf{Measure}}& $\rho^{\mathtt{MF}}$ & $\rho^{\mathtt{MP}}$	& $\rho^{\mathtt{MF}}$ & $\rho^{\mathtt{MP}}$	& $\rho^{\mathtt{MF}}$ & $\rho^{\mathtt{MP}}$\\\hline
				Variance & $0.0201$ & $\boldsymbol{0.0309}$ & $0.4593$ & $\boldsymbol{0.9366}$ & $0.0012$ & $\boldsymbol{0.0095}$\\
				Span &  $0.9567$ & $\boldsymbol{1.0387}$& $2.0000$ & $\boldsymbol{2.0000}$ & $0.4286$ & $\boldsymbol{2.0000}$  \\\midrule
				\textbf{Data}  & \multicolumn{2}{c}{$\makecell[c]{\boldsymbol{\mathrm{brakes\_abru\text{-}}}\\\boldsymbol{\mathrm{ptly\_relative}}}$} & \multicolumn{2}{c}{$\makecell[c]{\boldsymbol{\mathrm{keeps\_lane\_}}\\\boldsymbol{\mathrm{speed\_limit}}}$} & \multicolumn{2}{c}{$\makecell[c]{\boldsymbol{\mathrm{keeps\_fov\_}}\\\boldsymbol{\mathrm{speed\_limit}}}$} \\\hline  
				\noalign{\vspace{0.2ex}}${\bar{P}_{+,\min}}$ & \multicolumn{2}{c}{$0.3603$} & \multicolumn{2}{c}{$0.4942$} & \multicolumn{2}{c}{$0.0000$}\\
				${\bar{P}_{+,\max}}$ & \multicolumn{2}{c}{$0.4862$} & \multicolumn{2}{c}{$1.0000$} & \multicolumn{2}{c}{$1.0000$}\\ 
				${\bar{P}_{-,\min}}$ & \multicolumn{2}{c}{$0.1616$} & \multicolumn{2}{c}{$0.0646$} & \multicolumn{2}{c}{$0.0000$}\\
				${\bar{P}_{-,\max}}$ & \multicolumn{2}{c}{$0.6862$} & \multicolumn{2}{c}{$0.5030$} & \multicolumn{2}{c}{$1.0000$}\\
				Comp. Time & \multicolumn{2}{c}{$4.00 ms$} & \multicolumn{2}{c}{$3.96 ms$} & \multicolumn{2}{c}{$3.92 ms$}\\		
				Precision & \multicolumn{2}{c}{$1.0000$}& \multicolumn{2}{c}{$1.0000$} & \multicolumn{2}{c}{$1.0000$}\\
				Recall & \multicolumn{2}{c}{$1.0000$} & \multicolumn{2}{c}{$1.0000$} & \multicolumn{2}{c}{$1.0000$}\\
				F1-score & \multicolumn{2}{c}{$1.0000$}& \multicolumn{2}{c}{$1.0000$} & \multicolumn{2}{c}{$1.0000$}\\\hline {\textbf{Measure}}& $\rho^{\mathtt{MF}}$ & $\rho^{\mathtt{MP}}$	& $\rho^{\mathtt{MF}}$ & $\rho^{\mathtt{MP}}$	& $\rho^{\mathtt{MF}}$ & $\rho^{\mathtt{MP}}$\\\hline
				Variance & $0.0017$ & $\boldsymbol{0.0023}$ & $0.1412$ & $\boldsymbol{0.3789}$ & $0.0000$ & $\boldsymbol{0.0004}$ \\
				Span & $0.2381$ & $\boldsymbol{0.3226}$ & $1.1181$ & $\boldsymbol{1.9534}$ & $0.0000$ & $\boldsymbol{0.2301}$  \\\midrule
				\textbf{Data}  & \multicolumn{2}{c}{$\makecell[c]{\boldsymbol{\mathrm{keeps\_type\_}}\\\boldsymbol{\mathrm{speed\_limit}}}$} & \multicolumn{2}{c}{$\makecell[c]{\boldsymbol{\mathrm{keeps\_brake\_}}\\\boldsymbol{\mathrm{speed\_limit}}}$} & \\\cline{1-5}
				\noalign{\vspace{0.3ex}}${\bar{P}_{+,\min}}$ & \multicolumn{2}{c}{$0.0000$} & \multicolumn{2}{c}{$0.6973$} & \\
				${\bar{P}_{+,\max}}$ & \multicolumn{2}{c}{$1.0000$} & \multicolumn{2}{c}{$1.0000$} & \\ 
				${\bar{P}_{-,\min}}$ & \multicolumn{2}{c}{$0.0000$} & \multicolumn{2}{c}{$0.1052$} & \\
				${\bar{P}_{-,\max}}$ & \multicolumn{2}{c}{$1.0000$} & \multicolumn{2}{c}{$0.4558$} & \\
				Comp. Time & \multicolumn{2}{c}{$3.94 ms$} & \multicolumn{2}{c}{$3.96 ms$} &\\		
				Precision & \multicolumn{2}{c}{$1.0000$} & \multicolumn{2}{c}{$1.0000$} &\\
				Recall & \multicolumn{2}{c}{$1.0000$} & \multicolumn{2}{c}{$1.0000$} &\\
				F1-score & \multicolumn{2}{c}{$1.0000$} & \multicolumn{2}{c}{$1.0000$} &\\\cline{1-5} 
			{\textbf{Measure}}& $\rho^{\mathtt{MF}}$ & $\rho^{\mathtt{MP}}$	& $\rho^{\mathtt{MF}}$ & $\rho^{\mathtt{MP}}$\\\cline{1-5}
				Variance & $0.0000$ & $\boldsymbol{0.0026}$& $0.0000$ & $\boldsymbol{0.0017}$ &  & \\
				Span & $0.0000$ & $\boldsymbol{0.6706}$ & $0.0000$ & $\boldsymbol{1.0000}$ &  &  \\\cmidrule[1.pt]{1-5}
			\end{tabular}
		\end{center}
	\vspace{-8mm}
	\end{table}
	All computations were executed using a single thread on a machine equipped with two AMD EPYC 7763 64-Core processors and 2TB RAM. The detailed evaluation results are summarized in Tab.~\ref{tab:appendix}. {By evaluating the training set, we derive the normalization values in (\ref{eq:mb-robustness}). Notably, the model predictive robustness of each predicate can be efficiently computed, with an average processing time of less than $5 ms$.} Moreover, we assess the performance of the GP regression model by examining its ability to correctly classify the satisfaction or violation of predicates on the test set. The precision, recall, and F1-score for the classification all have an average value of $0.9999$. We can see that our proposed method maintains a high level of accuracy in robustness prediction, even without the rectification in (\ref{eq:clip}).}

\subsection{Comparison with Related Work}\label{sec:comp}
As discussed in Sec.~\ref{subsec:fea_var},
we consider comprehensive feature variables as inputs for the robustness prediction and can assess their predictive relevance utilizing the GP models. 
In contrast, the use of handcrafted functions to compute the robustness might lead to the lack of considered variables for complex predicates, which we demonstrate in the following example:

\vspace{1mm}
\noindent\textbf{Example}: Consider the predicate $\mathrm{in\_same\_lane}$ \cite{Maierhofer2020a} which describes whether the ego vehicle shares a lane with the vehicle $b\in\mathcal{B}$:
\begin{equation}\label{eq:in_same_lane}
	\mathrm{in\_same\_lane} \coloneqq |\mathcal{L}_{\text{ego}} \cap  \mathcal{L}_{b}| > 0.
\end{equation}
	The calculation of its model-free robustness in \cite[(1)]{ruleSTL} is sophisticated, but it only takes into account the positional attributes, represented as the signed distance to the lanes occupied by other vehicles. {Domain-specific normalization constants are then used to confine the robustness within the interval $[-1, 1]$ \cite[Tab.~III]{ruleSTL}. However, as depicted in Fig.~\ref{fig:compare_01}, the model-free robustness remains close to zero after normalization. {Instead, the model predictive robustness demonstrates greater variance and a wider span across all evaluated predicates, as can be observed in the measures provided in Tab.~\ref{tab:appendix}. This arises because the pre-defined normalization constants must be sufficiently large to account for all edge cases, which might rarely happen in the dataset.  For example, for the predicate $\mathrm{in\_same\_lane}$, a large normalization constant is selected by \cite{ruleSTL} to accommodate scenarios where vehicles are positioned significantly apart from each other laterally.}
Moreover, we use the variance of the GP posterior latent mean \cite{paananen2019variable} to analyze the sensitivity of the feature variables. 
As shown in Fig.~\ref{fig:comparison_rob}, the distribution reveals that not only the relative  distance to the lane bounds but also the relative velocity has a significant impact on the feature relevance. This is consistent with the human intuition of factors affecting the robustness of two vehicles maintaining the same lane, such as the variables used for calculating time-to-line-crossing \cite[Tab.~I]{mammar2006time}. {Additionally, this underscores the lack of monotonicity in the model-free robustness (cf. Prop.~\ref{prop:Monotonicity}), wherein a change in the relative velocity does not alter its value.}

\begin{figure}[t!]	
	\captionsetup[subfigure]{aboveskip=+1mm,belowskip=+1mm}
	\vspace{2.5mm}
	\centering
	\hspace{-4mm}\begin{subfigure}[b]{0.52\linewidth}
		\centering
		\scriptsize	
		\def\svgwidth{0.85
			\columnwidth}
\begingroup%
  \makeatletter%
  \providecommand\color[2][]{%
    \errmessage{(Inkscape) Color is used for the text in Inkscape, but the package 'color.sty' is not loaded}%
    \renewcommand\color[2][]{}%
  }%
  \providecommand\transparent[1]{%
    \errmessage{(Inkscape) Transparency is used (non-zero) for the text in Inkscape, but the package 'transparent.sty' is not loaded}%
    \renewcommand\transparent[1]{}%
  }%
  \providecommand\rotatebox[2]{#2}%
  \newcommand*\fsize{\dimexpr\f@size pt\relax}%
  \newcommand*\lineheight[1]{\fontsize{\fsize}{#1\fsize}\selectfont}%
  \ifx\svgwidth\undefined%
    \setlength{\unitlength}{407.73247433bp}%
    \ifx\svgscale\undefined%
      \relax%
    \else%
      \setlength{\unitlength}{\unitlength * \real{\svgscale}}%
    \fi%
  \else%
    \setlength{\unitlength}{\svgwidth}%
  \fi%
  \global\let\svgwidth\undefined%
  \global\let\svgscale\undefined%
  \makeatother%
  \begin{picture}(1,0.67065943)%
    \lineheight{1}%
    \setlength\tabcolsep{0pt}%
    \put(0,0){\includegraphics[width=\unitlength,page=1]{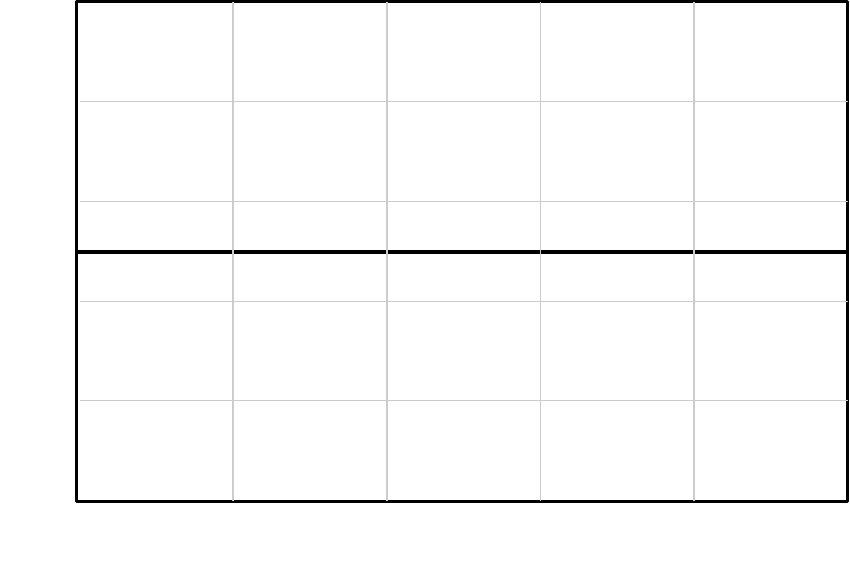}}%
    \put(0.04366507,0.64634599){\color[rgb]{0,0,0}\makebox(0,0)[lt]{\lineheight{1.25}\smash{\begin{tabular}[t]{l}$1$\end{tabular}}}}%
    \put(-0.0022961,0.06779237){\color[rgb]{0,0,0}\makebox(0,0)[lt]{\lineheight{1.25}\smash{\begin{tabular}[t]{l}$-1$\end{tabular}}}}%
    \put(0.06040075,0.00422575){\color[rgb]{0,0,0}\makebox(0,0)[lt]{\lineheight{1.25}\smash{\begin{tabular}[t]{l}$-1$\end{tabular}}}}%
    \put(0.98147614,0.00406545){\color[rgb]{0,0,0}\makebox(0,0)[lt]{\lineheight{1.25}\smash{\begin{tabular}[t]{l}$1$\end{tabular}}}}%
    \put(0.04777931,0.19015139){\color[rgb]{0,0,0}\rotatebox{90}{\makebox(0,0)[lt]{\lineheight{1.25}\smash{\begin{tabular}[t]{l}$\rho^{\mathtt{MP}}_\mathrm{in\_same\_lane}$\end{tabular}}}}}%
    \put(0.35520799,0.00942201){\color[rgb]{0,0,0}\makebox(0,0)[lt]{\lineheight{1.25}\smash{\begin{tabular}[t]{l}$\rho^{\mathtt{MF}}_{\mathrm{in\_same\_lane}}$\end{tabular}}}}%
    \put(0,0){\includegraphics[width=\unitlength,page=2]{compare_in_same_lane4.pdf}}%
  \end{picture}%
\endgroup%

		\caption{{\footnotesize{Robustness comparison. 
			}}
		}\label{fig:compare_01}
	\end{subfigure}\hspace{-2mm}
	\begin{subfigure}[b]{0.515\linewidth}
		\centering
		\scriptsize
		\def\svgwidth{1.\columnwidth}
		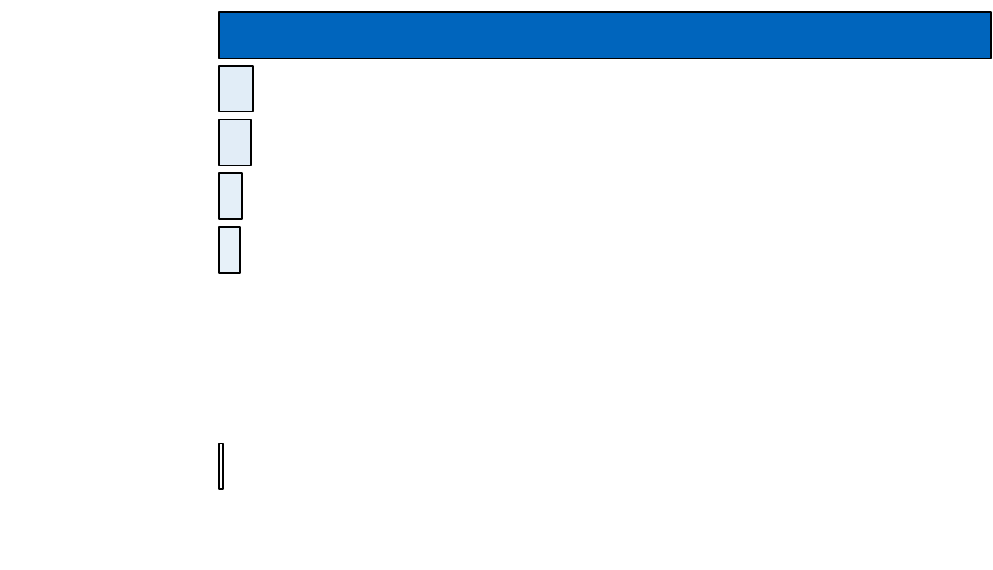
		\caption{{\footnotesize Sensitivity analysis.}}\label{fig:comparison_rob}
	\end{subfigure}	
	\caption{Evaluation results of the predicate $\mathrm{in\_same\_lane}$. (a) shows the distribution of its model-free and model predictive robustness with $1,000$ randomly selected data points, while (b) analyzes the sensitivity  of the feature variables, where the values are scaled so that the most relevant feature variable has a relevance of one. }\label{fig:compare-in-same-lane}
	\vspace{-4mm}
\end{figure}
\begin{figure*}[!htbp]
	\captionsetup[subfigure]{aboveskip=+1mm,belowskip=+1mm}
	\centering
	\vspace{2.5mm}
	\begin{subfigure}[b]{1\linewidth}
		\footnotesize
		\centering
		\def\svgwidth{0.99\columnwidth}
		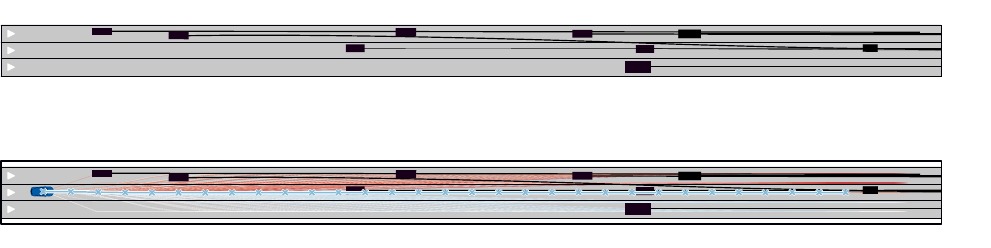\vspace{0mm}
		\caption{{\footnotesize Scenario configuration and planning results.}}\label{fig:rob-mp-a}
		\vspace{2mm}
	\end{subfigure}
	\begin{subfigure}[b]{0.49\linewidth}
		\footnotesize
		\centering
		\def\svgwidth{0.99\columnwidth}
		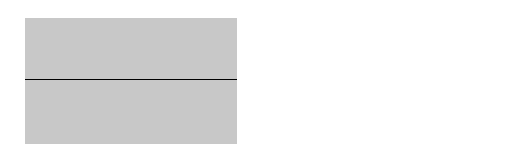
		\caption{{\footnotesize Model-free robustness.}}\label{fig:rob-mp-b}
	\end{subfigure}
	\begin{subfigure}[b]{0.49\linewidth}
	\footnotesize
	\centering
	\def\svgwidth{0.99\columnwidth}
	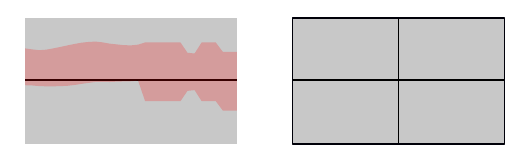
	\caption{{\footnotesize Model predictive robustness.}}\label{fig:rob-mp-b2}
\end{subfigure}\vspace{0.0mm}	
	\caption{\footnotesize Robustness-aware trajectory planning. 
	The trajectories in (a) are color-coded according to the robustness, which increases from red with negative to blue with positive values. {(b) and (c) depict the evaluation results from using model-free and model predictive robustness, respectively.} The shaded regions in (b) denote the $2$-$\sigma$ model uncertainty corresponding to a 95.4\% confidence level.  In the right-hand panels of (b) and (c), the blue dots to the right of the yellow dotted line indicate that the robustness of the optimal trajectory is higher than that of the human trajectory. The red dots to the left of the line indicate a lower robustness. }\label{fig:rob-mp}
\vspace{-4mm}
\end{figure*}
\subsection{Robustness-aware Trajectory Planning}
Our robustness measure can be easily integrated into the prediction of traffic rule violations \cite{ruleSTL} and trajectory repairing \cite{YuanfeiLin2022a}. In this section, we demonstrate that the model-predictive definition also facilitates the robustness awareness of trajectory planning using a sampling-based planner of~\cite{werling2012optimal}. The robustness of rules $\varphi_{\text{R\_G1}}$ to $\varphi_{\text{R\_G3}}$\footnote{$\varphi_{\text{R\_G1}}$: Safe distance to preceding vehicle; $\varphi_{\text{R\_G2}}$: Unnecessary braking; $\varphi_{\text{R\_G3}}$: Maximum speed limit.} from \cite{Maierhofer2020a} is integrated as an additional robustness term $J_r$ with weight $\lambda_r\in\mathbb{R}_{>0}$ in the cost function $J$ to the planner:
\begin{equation*}
	J(x_{\text{ego}, k}, u_{\text{ego}, k}) = J_p(x_{\text{ego}, k}, u_{\text{ego}, k}) \!-\! \lambda_r J_r(x_{\text{ego}, k}, u_{\text{ego}, k}),
\end{equation*}
where the performance term $J_p$ is obtained from \cite[(4)]{werling2012optimal} and the robustness calculation of STL operators is based on \cite[Def. 3]{donze2010robust}. {The robustness term $J_r$ is denoted as $J_r^{\mathtt{MF}}$ and $J_r^{\mathtt{MP}}$ when using model-free and model predictive robustness, respectively, for the predicate evaluation, and is defined as:}
\begin{equation*}
	\begin{split}
		 J_r^{\mathtt{MP}}(x_{\text{ego}, k}, u_{\text{ego}, k}) &\coloneqq  \tilde{\rho}^{\mathtt{MP}}_{\varphi_{\text{R\_G1}}\wedge\varphi_{\text{R\_G2}}\wedge\varphi_{\text{R\_G3}}}(\boldsymbol{\signal}, k),\\
		 J_r^{\mathtt{MF}}(x_{\text{ego}, k}, u_{\text{ego}, k}) &\coloneqq  {\rho}^{\mathtt{MF}}_{\varphi_{\text{R\_G1}}\wedge\varphi_{\text{R\_G2}}\wedge\varphi_{\text{R\_G3}}}(\boldsymbol{\signal}, k).
	\end{split}
\end{equation*}
During planning, the collision-free sample with the minimum cost is selected as the optimal trajectory. 

We show an exemplary scenario in Fig.~\ref{fig:rob-mp-a}, where 450 trajectories are generated with a time increment of $0.2s$ and a horizon of $30$ time steps. The robustness distribution of the selected optimal trajectory is compared to the one of the recorded human trajectory in the left panel of Fig.~\ref{fig:rob-mp-b} and Fig.~\ref{fig:rob-mp-b2}. 
{As the robustness of the human trajectory falls below $0$ for both evaluations at $k\in[28, 30]$, which are sound (cf. Tab.~\ref{tab:predicates}), it indicates a violation of the traffic rules by the human driver.
In particular, this violation is due to the driver failing to maintain a safe distance from the vehicle ahead.
 By incorporating robustness, either model-free or model predictive, into the cost function, the ego vehicle brakes in such a way as to obey traffic rules as much as possible while minimizing energy consumption. However, as the model-free definition is less comprehensive (cf.~Sec.\ref{sec:comp})
 , it tends to result in harder braking for the ego vehicle, as observed in the optimal trajectories in Fig.~\ref{fig:rob-mp-a}.} 

Furthermore, we evaluate the robustness-aware planning algorithm on 100 randomly selected highD scenarios. {As shown in the right panel of Fig.~\ref{fig:rob-mp-b}, $53.95\%$ of the planned trajectories have greater robustness compared to the recorded ones using the model-free robustness. The average increment of the model-free robustness is $48.08\%$. In contrast, using the model predictive robustness results in a similar average robustness increment of $51.84\%,$ but a substantial increase to ${88.23\%}$ of the optimal trajectories demonstrating higher robustness than those of the human driver (cf. the right panel of Fig.~\ref{fig:rob-mp-b2}).} The results show that the \RobName robustness helps to enhance the degree of traffic rule compliance of autonomous vehicles, which significantly outperforms human drivers {and the model-free approach} in the evaluated dataset. {Due to the systematic nature of our definition, incorporating additional rules with new predicates into the planner is straightforward.} 
\section{Conclusions}\label{sec:conc}
This paper examines how one can precisely and quickly quantify the level of  satisfaction or violation of STL predicates considering the environment and dynamic models.
Unlike existing model-free robustness definitions, our proposed \RobName robustness not only grows monotonically as the satisfaction probability increases but also is sound and smooth in the sense of Prop. \ref{prop:soundness}-\ref{prop:Monotonicity}. With this, our method can be useful in terms of rule-compliant planning and control for autonomous vehicles. 
{It is important to note that this work primarily showcases the benefits of our approach for predicates designed for German interstates. To apply our approach to other predicates and datasets, careful design of the prediction model and feature variables may be necessary.  

%
\IEEEpeerreviewmaketitle

\section*{Acknowledgment}
The authors kindly thank Patrick Halder for his valuable suggestions on earlier drafts of this paper and Ethan Tatlow for the voice-over in the video attachment.

\ifCLASSOPTIONcaptionsoff
  \newpage
\fi

\bibliographystyle{IEEEtran}
\balance{\bibliography{robustness.bib, summary.bib}}

\end{document}